%% file: main.tex
\documentclass[11pt]{article}

\usepackage[preprint]{acl}

\usepackage{times}
\usepackage{latexsym}

\usepackage[T1]{fontenc}

\usepackage[utf8]{inputenc}

\usepackage{microtype}

\usepackage{inconsolata}

\usepackage{graphicx}

\input{cmd}

\input{title}

\hypersetup{
  pdftitle={Constraint-Aware Synthetic Tabular Data Generation via Inter-Column Constraint Discovery with LLM Agents},
  pdfauthor={Jianxing Zhao, Mao Guan, Dongyu Liu}
}

\begin{document}
\raggedbottom
\maketitle
\begingroup
\renewcommand{\thefootnote}{\fnsymbol{footnote}}
\footnotetext[1]{Jianxing Zhao and Dongyu Liu are with the University of California, Davis.}
\footnotetext[2]{Mao Guan is an independent researcher.}
\endgroup
\footnotetext[1]{Code and data at \url{https://github.com/via-cs/dive-tabular}.}
\setcounter{footnote}{1}
\input{abstract}

\input{introduction}

\input{problem_setting}

\input{methodology}

\input{experiments}

\input{conclusion}

\input{limitations}
\input{ethics_statement}



\FloatBarrier
\bibliography{custom}

\clearpage
\appendix
\input{appendices}

\end{document}

%% file: cmd.tex
\usepackage{listings}
\usepackage[most]{tcolorbox}

\usepackage{algorithm}
\usepackage{algpseudocode}

\usepackage{amsmath}
\usepackage{amssymb}
\usepackage{array}
\usepackage{booktabs}
\usepackage{multirow}
\usepackage{placeins}

\definecolor{codeblue}{RGB}{36,99,161}
\definecolor{codegreen}{RGB}{40,130,80}
\definecolor{codegray}{RGB}{95,99,104}
\definecolor{codebackground}{RGB}{247,248,250}
\definecolor{codeborder}{RGB}{215,218,222}
\definecolor{eqaccent}{HTML}{6558C7}
\definecolor{eqbackground}{HTML}{F5F3FF}
\definecolor{fdaccent}{HTML}{2B6FAE}
\definecolor{fdbackground}{HTML}{F0F6FC}
\definecolor{cataccent}{HTML}{16857F}
\definecolor{catbackground}{HTML}{EFF9F7}
\definecolor{linaccent}{HTML}{A96813}
\definecolor{linbackground}{HTML}{FFF7E8}

\newcommand{\deltapos}[1]{\begingroup\setlength{\fboxsep}{0.8pt}%
    \colorbox{catbackground}{\textcolor{cataccent}{\bfseries #1}}\endgroup}
\newcommand{\deltaneg}[1]{\begingroup\setlength{\fboxsep}{0.8pt}%
    \colorbox{linbackground}{\textcolor{linaccent}{\bfseries #1}}\endgroup}
\newcommand{\deltazero}{\begingroup\setlength{\fboxsep}{0.8pt}%
    \colorbox{codebackground}{\textcolor{codegray}{\bfseries 0.000}}\endgroup}

\lstdefinestyle{python}{
    language=Python,
    basicstyle=\ttfamily\small,
    keywordstyle=\color{codeblue}\bfseries,
    stringstyle=\color{codegreen},
    commentstyle=\color{codegray}\itshape,
    backgroundcolor=\color{codebackground},
    frame=single,
    rulecolor=\color{codeborder},
    showstringspaces=false,
    breaklines=true,
    columns=fullflexible,
    keepspaces=true,
    xleftmargin=3pt,
    xrightmargin=3pt,
    aboveskip=5pt,
    belowskip=5pt
}

\lstdefinestyle{constraintdisplay}{
    basicstyle=\ttfamily\scriptsize,
    showstringspaces=false,
    breaklines=true,
    breakatwhitespace=false,
    columns=fullflexible,
    keepspaces=true,
    frame=none,
    xleftmargin=0pt,
    xrightmargin=0pt,
    aboveskip=0pt,
    belowskip=0pt
}

\lstdefinestyle{constraintpython}{
    style=constraintdisplay,
    language=Python,
    keywordstyle=\color{codeblue}\bfseries,
    stringstyle=\color{codegreen},
    commentstyle=\color{codegray}\itshape
}

\newcommand{\dataset}[1]{\textsc{\MakeLowercase{#1}}}

\newcounter{constraintexample}[section]
\renewcommand{\theconstraintexample}{%
    \thesection.\arabic{constraintexample}}

\tcbset{
    constraint card/.style={
        enhanced jigsaw,
        listing only,
        listing engine=listings,
        boxrule=0.45pt,
        arc=1.4mm,
        outer arc=1.4mm,
        left=1.2mm,
        right=1.2mm,
        top=1.1mm,
        bottom=1.1mm,
        boxsep=0.5mm,
        fonttitle=\sffamily\bfseries\small,
        coltitle=black,
        titlerule=0pt,
        before skip=8pt,
        after skip=9pt,
        listing options={style=constraintdisplay}
    }
}

\lstdefinestyle{constraintprompt}{
    style=constraintdisplay,
    basicstyle=\ttfamily\fontsize{8}{9.5}\selectfont,
    breakatwhitespace=true,
    breakautoindent=false,
    breakindent=0pt,
    keywordstyle={},
    stringstyle={},
    commentstyle={}
}

\definecolor{dypink}{HTML}{ec008c}
\definecolor{dypurple}{HTML}{8654d1}
\definecolor{jxgreen}{HTML}{228B22}

\newif\ifnotes
\notestrue 



%% file: title.tex
\title{Constraint-Aware Synthetic Tabular Data Generation via Inter-Column Constraint Discovery with LLM Agents}

\author{
  \textbf{Jianxing Zhao}\textsuperscript{*} \quad
  \textbf{Mao Guan}\textsuperscript{\textdagger} \quad
  \textbf{Dongyu Liu}\textsuperscript{*} \\
  \texttt{jxrzhao@ucdavis.edu} \quad
  \texttt{guanmao771@gmail.com} \quad
  \texttt{dyuliu@ucdavis.edu}
}


%% file: abstract.tex
\begin{abstract}
    Generating structurally valid synthetic tabular data remains difficult:
    outputs with high statistical fidelity and downstream utility can still violate
    semantically meaningful domain constraints. 
    We study the discovery and
    enforcement of three complementary inter-column constraint
    families---equations, linear inequalities, and logical dependencies. Our
    unified tool-grounded workflow represents all three as machine-executable
    hypotheses and applies a common interface for full-table validation,
    deterministic diagnosis, and counterexample-guided revision.
    A generator-agnostic postprocessor coordinates family-specific repairs
    on outputs from unchanged tabular generators. 
    Across curated behavioral
    audits and end-to-end evaluations, the complete workflow improves held-out
    violation detection over one-shot direct prompting, while postprocessing
    yields zero measured violations for every retained, applicable constraint,
    improves downstream utility on most datasets, and largely preserves
    univariate marginals.
\end{abstract}

%% file: introduction.tex
\section{Introduction}

Synthetic tabular data supports model development, system testing, and
analysis when real records are scarce or difficult to use
\citep{shi2025comprehensive}. Yet outputs with strong distributional similarity
and downstream utility can remain structurally implausible: individual records
may violate domain constraints
\citep{stoian2024how,umesh2025preserving,afonja2026from}. An e-commerce
generator may, for example, produce an order whose delivery precedes placement,
whose subtotal disagrees with its unit price and quantity, or whose delivery
method is incompatible with the product type. Such violations can undermine
simulation and decision support, distort downstream models, and make synthetic
records distinguishable from real ones through simple logical checks
\citep{long2025llm,afonja2026from,dhooghe2026moet}.

Inter-column constraints make structural validity explicit and testable.
Prior work either enforces supplied rules through generator-integrated
objectives, dependency-aware synthesis, or inference-time refinement
\citep{vero2023cuts,stoian2024how,stoian2025beyond,dhooghe2026moet}, or
automates acquisition within a restricted constraint family or synthesis
pipeline \citep{umesh2025preserving,afonja2026from,long2025llm}. These methods
do not jointly address two coupled challenges. First, heterogeneous semantic
relationships must be represented as executable hypotheses that can be tested
and revised against the complete reference table. Second, enforcement on
outputs from unchanged tabular generators must coordinate repairs across shared columns,
because one correction can invalidate another or unnecessarily distort the
generated distribution. Section~\ref{sec:problem_setting} reviews these lines
of work and formalizes our setting.

We address these gaps with an end-to-end framework for tool-grounded
discovery and generator-agnostic enforcement. LLM agents use metadata, column
profiles, and sampled records to propose equations, linear inequalities, and
logical dependencies in machine-executable forms. Family-specific deterministic
tools test every proposal against the complete reference table and return
diagnostics and counterexamples for iterative revision. An ordered
postprocessor then enforces the retained constraints on outputs from
unchanged tabular generators, using distribution-aware repairs while protecting
relationships restored earlier. The central contribution is this shared
executable interface and coordinated postprocessor: LLM outputs are treated as
revisable hypotheses and heterogeneous constraints as an interacting system.

We evaluate discovery on three curated benchmarks with two LLM backbones
and end-to-end constraint-aware generation on seven public datasets across four
generator paradigms. On held-out contrastive audit tasks, the
complete executable validation-and-revision workflow achieves higher mean
violation-detection accuracy, precision, and recall than one-shot direct prompting in all
18 family--backbone--metric comparisons. Postprocessing yields zero
measured violations under every retained, applicable validator; mean TSTR
utility improves on four datasets and declines by at most 0.042 on the other
three, while mean changes in Column Shapes range from $-0.009$ to $+0.019$.

Our contributions are:
\begin{enumerate}
    \setlength{\itemsep}{0pt}
    \setlength{\parsep}{0pt}
    \setlength{\parskip}{0pt}
    \item A unified, typed machine-executable interface for equational,
    linear-inequality, and logical-dependency hypotheses, supporting full-table
    validation, deterministic diagnosis, and counterexample-guided revision.
    \item A generator-agnostic postprocessor that coordinates categorical
    and numerical repairs across shared columns without backbone access or
    retraining.
    \item Empirical validation across three discovery benchmarks, two LLM
    backbones, seven end-to-end datasets, and four generator paradigms,
    demonstrating improved held-out violation detection, zero measured
    violations under retained, applicable constraints, and largely preserved
    utility and univariate marginal fidelity.
\end{enumerate}

%% file: problem_setting.tex
\newcommand{\colname}[1]{\texttt{\detokenize{#1}}}

\section{\texorpdfstring{Related Work and Problem Setting}
{Related Work and Problem Setting}}\label{sec:problem_setting}

\subsection{Tabular Synthetic Data Generation}\label{sec:tabular_generation}

Let $\mathcal{D}_{\mathrm{real}}=\{\mathbf{x}_i\}_{i=1}^{N}$ be a real
tabular dataset with $d$ columns. The domain of column $j$ is denoted by
$\mathcal{X}_j$ and may be numerical, with
$\mathcal{X}_j\subseteq\mathbb{R}$, or categorical, with
$\mathcal{X}_j$ a finite discrete set. Each row (record)
$\mathbf{x}_i=(x_{i1},\ldots,x_{id})$  therefore belongs to the heterogeneous
product domain $\mathcal{X}=\prod_{j=1}^{d}\mathcal{X}_j$ and may contain
both continuous and discrete values. 
We treat the rows as i.i.d. samples
from an underlying mixed-type joint distribution $p_{\mathrm{data}}$ on
$\mathcal{X}$.
Let $\mathcal{M}$ contain the available metadata, such as the dataset
description, column names, data types, and column descriptions.

A tabular generator learns a distribution $p_{\theta}$ over $\mathcal{X}$
from $\mathcal{D}_{\mathrm{real}}$ to approximate $p_{\mathrm{data}}$ and
produces a synthetic dataset
\[
    \mathcal{D}_{\mathrm{syn}}
    =
    \{\mathbf{x}^{\mathrm{syn}}_i\}_{i=1}^{M},
    \qquad
    \mathbf{x}^{\mathrm{syn}}_i
    \overset{\mathrm{i.i.d.}}{\sim}
    p_{\theta}.
\]
Methods for learning $p_{\theta}$ span copula-based statistical models
\citep{patki2016synthetic}; deep generators including CTGAN and TVAE
\citep{xu2019modeling}, TableGAN \citep{park2018data}, GOGGLE
\citep{liu2023goggle}, TabDDPM \citep{kotelnikov2023tabddpm}, and STaSy
\citep{kim2023stasy}; and language-model generators. GReaT and TapTap learn
from serialized rows through fine-tuning or pretraining
\citep{DBLP:conf/iclr/BorisovSLPK23,DBLP:conf/emnlp/ZhangWYJL23}, whereas AIGT,
EPIC, and Curated LLM use prompting or LLM-guided curation
\citep{zhang-etal-2025-aigt,DBLP:conf/nips/KimKC24,DBLP:conf/icml/SeedatHBS24}.
Recent surveys provide broader coverage
\citep{shi2025comprehensive,stoian2025survey}. Across paradigms, evaluation
primarily emphasizes marginal and inter-column distributional similarity and
downstream utility, but these aggregate criteria cannot ensure that every
record obeys domain semantics \citep{umesh2025preserving}.

\subsection{Constraint-Aware Generation}
\label{sec:constraint_motivation}


We study constraint-aware generation, in which a synthetic table should
both approximate the real-data distribution and satisfy inter-column
constraints implied by dataset semantics and metadata. Prior systems
incorporate user-specified logical and statistical rules through differentiable
objectives and rejection sampling \citep{vero2023cuts}; enforce supplied
linear-arithmetic constraints during training or inference-time refinement
\citep{stoian2024how,stoian2025beyond}; or reconstruct dependent features from
supplied mappings, dependency graphs, and derivation functions
\citep{umesh2026dependency,dhooghe2026moet}.

We define an instantiated inter-column constraint as a row-separable
predicate $c:\mathcal{X}\rightarrow\{0,1\}$ over two or more columns, where
$c(\mathbf{x})=1$ denotes satisfaction. It is evaluated independently on each
record and never compares different records. Our record-level scope is
narrower than that of prior systems: it excludes uniqueness and cross-table key
constraints supported by SDV \citep{datacebo2026sdv}, as well as dataset-level
statistical constraints considered by CuTS \citep{vero2023cuts}.

We focus on three complementary, rather than exhaustive,
constraint families: equations capture numerical derivation, linear
inequalities numerical feasibility, and logical dependencies categorical
admissibility. Together they span numerical and categorical columns and
deterministic and set-valued semantics while remaining executable and
verifiable at the record level. We use a hypothetical e-commerce order
table as a running example.

\textbf{Equational constraints} capture deterministic arithmetic
relationships among numerical columns. In the running example, the identity
$\colname{subtotal}=\colname{unit_price}\times\colname{quantity}$ states that
the subtotal is determined exactly by the unit price and quantity. Our
executable checkers support well-defined arithmetic identities without
requiring linearity, convexity, or differentiability; related dependency-aware
generation encodes such identities through supplied derivation functions
\citep{dhooghe2026moet}.

\textbf{Linear inequality constraints} define feasible regions
over numerical columns, expressing relative bounds, orderings, and capacity
limits. For the same table,
$\colname{delivery_time}\geq\colname{order_time}$ requires delivery to occur no
earlier than order placement. Unlike an equation, an inequality generally
permits multiple valid assignments, while its convex structure enables
tractable joint projection \citep{stoian2024how,afonja2026from}.

\textbf{Logical dependency (LD) constraints} restrict categorical
co-occurrences: for each configuration of determinant columns, the dependent
column may take only a specified set of admissible values
\citep{umesh2025preserving}. In this example,
$\colname{zipcode}\rightarrow\colname{state}$ is an FD: each ZIP code
determines one state. A partial rule applies when
$\colname{type}=\colname{digital}$ and $\colname{status}=\colname{paid}$,
restricting $\colname{delivery}$ to
$\{\colname{email},\colname{download},\colname{in-app}\}$ while permitting
multiple valid channels. An FD is a special
case of an LD in which each determinant configuration maps to
exactly one dependent value;
We adopt a broader definition of LD that permits multiple admissible
values and applies constraints only to selected determinant configurations,
thereby capturing partial and set-valued categorical rules
\citep{umesh2025preserving,umesh2026dependency,dhooghe2026moet}.

These families require distinct validation and repair operators:
equations require tolerance-aware checking and arithmetic reconstruction,
inequalities require joint feasibility and numerical projection, and LDs require discrete admissibility maps and conditional replacement. Moreover,
constraints can share columns, so repairing one can invalidate another or
unnecessarily distort the generated distribution; joint enforcement therefore
requires a coordinated repair order rather than independent corrections.

\subsection{Constraint Discovery and Enforcement}
\label{sec:constraint_objective}

Automated constraint acquisition remains fragmented. Data-driven methods
mine functional or logical dependencies \citep{umesh2025preserving}; LLM-based
systems infer graphical or logical structure within prompt-based or
latent-diffusion synthesis pipelines
\citep{liu2025structsynth,long2025llm}; and other workflows generate executable
data-quality rules or propose, filter, and select linear inequalities
\citep{akella-etal-2025-quality,afonja2026from}. These methods address
individual constraint families or couple acquisition to a particular synthesis
pipeline.

Given $\mathcal{D}_{\mathrm{real}}$, its metadata $\mathcal{M}$, and an
unconstrained synthetic table $\mathcal{D}_{\mathrm{syn}}$, our first objective
is to discover executable equational, linear-inequality, and
logical-dependency constraints, validate them against all reference rows, and
revise failed hypotheses using counterexamples. Our second objective is to
postprocess $\mathcal{D}_{\mathrm{syn}}$ so that these constraints are jointly
satisfied while preserving statistical fidelity and downstream utility.
Existing approaches provide subsets of these capabilities---for example,
P-DGM refines black-box outputs under supplied linear constraints
\citep{stoian2024how}. Our framework instead couples a shared executable
discovery-and-revision interface across three families with coordinated
post-hoc repair, without generator access or retraining.

%% file: methodology.tex
\section{Methodology}
\label{sec:methodology}

\begin{figure}[t]
  \centering
  \includegraphics[width=\columnwidth]{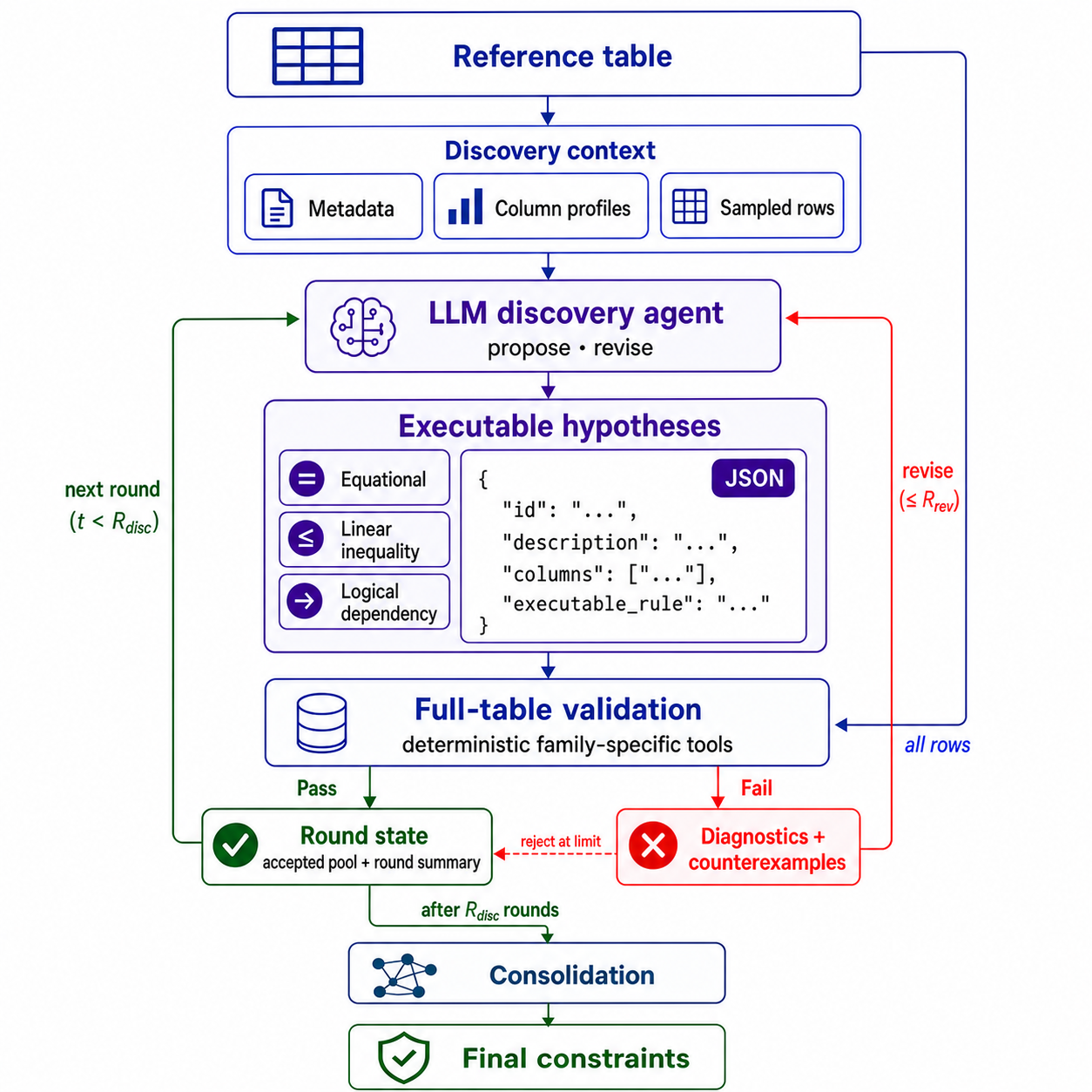}
  \caption{Overview of the LLM-guided constraint discovery workflow.}
  \label{fig:overall_framework}
\end{figure}

\subsection{LLM-Guided Constraint Discovery}
\label{sec:constraint_discovery}

As summarized in Figure~\ref{fig:overall_framework}, our family-specific
LLM workflow profiles the reference table, generates executable hypotheses,
validates and revises them using deterministic tools, and consolidates accepted
constraints for subsequent enforcement.

\paragraph{Dataset profiling.}
We first construct compact column profiles of $\mathcal{D}_{\mathrm{real}}$, similar to the practices in 
previous works involving LLMs in tabular data analysis \citep{10.1145/3772318.3791857,long2025llm,akella-etal-2025-quality}. 
For each column, the profile contains its semantic description, data types (num/cat),  
and type-specific statistics. Numerical
statistics include ranges, quantiles, means, stds, and mass points, while categorical statistics
include frequent 
values 
and their frequencies.

\paragraph{Executable constraint hypothesis generation.}
For each constraint family, the discovery LLM agent receives the dataset description, column profiles, 
and a context set of $n_{\mathrm{ctx}} \ll N$ reference rows sampled without replacement.
Using a fixed cap keeps prompt length and inference cost bounded. 
The sampled rows are used only for hypothesis generation; all generated hypotheses are validated 
against the full table.

The agent emits each candidate as a structured hypothesis containing (1) a
unique identifier, (2) a natural-language description, (3) the participating
columns, and (4) a family-specific machine-executable representation.
The executable field is central to our design: it converts a semantic
proposal into a deterministically evaluable object and provides a common
interface for proposal, full-table execution, diagnosis, and revision across
all three constraint families.
The description supports human inspection, but only the executable field is
used for validation. Prior work has generated executable tabular-quality
validators or directly parseable linear constraints
\citep{akella-etal-2025-quality,afonja2026from}; our framework unifies these
capabilities within a typed validation-and-revision workflow.
Appendix~\ref{app:constraint_representations} gives concrete examples of
all supported representations.

An equational candidate $h$ uses executable Python to represent an
arithmetic identity among numerical columns. The generated code defines a
single \texttt{check(df)} function that evaluates the identity row by row and
returns an index-aligned Boolean Series. Because equations may involve
free-form arithmetic expressions and dataset-specific numerical tolerances,
an executable Python checker is more expressive than a fixed-vocabulary 
domain-specific language (DSL) while
remaining deterministically evaluable
(Example~\ref{ex:eq-hypothesis}).

%
%

%
%

A linear-inequality candidate $h$ uses a JSON-encoded DSL to represent the
feasible half-space
\[
    h(\mathbf{x}) :
    \sum_{j\in S_h} a_{h,j}x_j
    \mathrel{\bowtie_h} b_h,
    \qquad
    \bowtie_h\in\{\leq,\geq\},
\]
where $S_h\subseteq\{1,\ldots,d\}$ indexes the participating numerical
columns,
$a_{h,j}$ is the coefficient of column $j$, and $b_h$ is the boundary
constant. The DSL stores the participating columns, coefficient map,
comparison sense, and right-hand-side constant, following prior work on
constraint extraction \citep{afonja2026from}. By explicitly encoding both the
boundary and its feasible side, this representation supports deterministic
parsing, evaluation, canonicalization, and joint projection
(Example~\ref{ex:linear-hypothesis}).


An LD candidate $h$ uses a unified JSON-encoded DSL to represent
categorical admissibility. It specifies one or more determinant columns
$\mathbf{X}_h=(X_{h,1},\ldots,X_{h,k_h})$, one dependent column $Y_h$, and a
value table $T_h$. Each entry of $T_h$ associates a nonempty value set for
every determinant column with a nonempty set of admissible dependent values.
A row matches an entry when its value for every determinant column belongs to
the corresponding set, and it satisfies the entry when its dependent value
belongs to the associated admissible set. Values within a determinant set are
alternatives, whereas sets across determinant columns are matched jointly.
This representation enables deterministic equality and set-membership tests
without imposing an arbitrary geometry on categorical codes
(Examples~\ref{ex:logical-value-table} and~\ref{ex:fd-shorthand}).

\paragraph{Tool-grounded validation and revision.}
Every proposed hypothesis $h$ is evaluated on the complete reference
dataset by a deterministic, family-specific tool:
\texttt{validate\_equational}, \texttt{validate\_linear}, or
\texttt{validate\_LD}. Let
$\mathcal{A}_h\subseteq\mathcal{D}_{\mathrm{real}}$ denote the rows to which
$h$ applies and $\mathcal{V}_h\subseteq\mathcal{A}_h$ the rows that violate it.
All three validators compute
$\operatorname{VR}(h)=|\mathcal{V}_h|/|\mathcal{A}_h|$, and retain $h$ only if
$\operatorname{VR}(h)\leq\tau_{\mathrm{vio}}$.
Equational and linear-inequality hypotheses apply to every row, so
$\mathcal{A}_h=\mathcal{D}_{\mathrm{real}}$. For an LD hypothesis,
$\mathcal{A}_h$ contains only rows whose determinant values match at least one
entry of its value table $T_h$; rows that match no entry are outside the
hypothesis's scope.

Each validator computes $|\mathcal{V}_h|$ from the family-specific executable
representation defined above. \texttt{validate\_equational} executes the
generated \texttt{check(df)} function and counts the \texttt{False} entries in
its returned Boolean Series. \texttt{validate\_linear} evaluates the DSL-defined
half-space and counts rows outside its feasible region. \texttt{validate\_LD}
counts applicable rows with dependent values outside their matched admissible
sets.

LD hypotheses additionally require sufficient empirical support.
Association-rule mining uses support to distinguish statistically meaningful
rules from patterns observed too infrequently to merit consideration
\citep{agrawal1993mining}. Analogously, \texttt{validate\_LD} calculates applicability support as
$\operatorname{Supp}(h)=|\mathcal{A}_h|/N$ and retain an LD hypothesis only if
$\operatorname{Supp}(h)\geq\tau_{\mathrm{sup}}$. This criterion complements
the violation rate by measuring the breadth of evidence for the hypothesis,
preventing a narrowly scoped rule from passing on too few applicable rows.
We set both thresholds to $0.005$ in all experiments, i.e.,
$\tau_{\mathrm{vio}}=\tau_{\mathrm{sup}}=0.005$.

When $h$ fails validation, the validator returns aggregate diagnostics and
up to $n_{\mathrm{cex}}$ violating rows as counterexamples to the same discovery agent
to request for a revision.
The agent may make at most $R_{\mathrm{rev}}$ revision attempts after the
initial validation, stopping early upon acceptance. A candidate that still fails after the final attempt is
rejected.

\paragraph{Multi-round discovery.}
A single discovery LLM call may focus on the most salient relationships and overlook
less obvious constraints. We therefore conduct $R_{\mathrm{disc}}$
discovery rounds per constraint family for every dataset.
In each follow-up round, the agent receives a summary of the hypotheses
accepted and rejected in previous rounds and is instructed to explore 
explore relationships that haven't been explored in previous discovery rounds.
Unless otherwise stated, we use $n_{\mathrm{ctx}}=100$ context rows,
$n_{\mathrm{cex}}=20$ counterexamples, $R_{\mathrm{rev}}=3$ revision attempts,
and $R_{\mathrm{disc}}=3$ discovery rounds. This configuration balances broad
constraint coverage with token usage, as shown by the ablation of
$n_{\mathrm{ctx}}$, $n_{\mathrm{cex}}$, and $R_{\mathrm{disc}}$ in
Appendix~\ref{app:discovery_hyperparameter_sensitivity}.




\paragraph{Consolidation.}
After $R_{\mathrm{disc}}$ discovery rounds, we apply family-specific consolidation to remove
redundant or structurally overlapping constraints.
For linear inequalities, we first rewrite each candidate as
$\mathbf{a}_h^\top\mathbf{x}\leq b_h$ and normalize it by a positive scale.
We remove exact duplicates and, among inequalities with an identical
normalized coefficient vector $\mathbf{a}_h$, retain only the smallest
right-hand side $b_h$.
For equational constraints, we first retain one preferred candidate for each participating-column 
set. We then solve a binary optimization problem that maximizes the number of retained equations
subject to every retained equation having at least one participating column unused by every other 
retained equation. For LD constraints, we enforce an acyclic dependency graph in acceptance order, 
canonicalize value tables into atomic mappings, remove exact duplicates and strictly subsumed mappings, 
and merge compatible mappings with the same determinant–dependent signature into one value table representation.

\subsection{Constraint Enforcement}
\label{sec:constraint_enforcement}
We postprocess $\mathcal{D}_{\mathrm{syn}}$ without accessing or changing
the backbone, making the same pipeline applicable to black-box generators and
already-produced tables. This avoids the retraining or
training-pipeline changes required by generator-integrated methods
\citep{stoian2024how,vero2023cuts,afonja2026from,long2025llm}.
We also avoid rejection sampling: at high violation rates, acceptance
becomes impractically low, while selective retention can distort the generated
distribution and downstream utility \citep{stoian2025beyond}, which motivates
direct post-hoc constraint repair.

We enforce logical dependencies, equations, and linear inequalities in sequence. 
Because logical repairs modify only categorical columns, their ordering relative 
to the numerical repair stages is immaterial. Equation repair, however, must strictly 
precede linear projection: reconstructing a column after projection can reintroduce 
inequality violations. During projection, we therefore hold equation-participating columns 
fixed, preserving the repaired equations throughout the final stage. 
Our ablation in Appendix~\ref{app:cross_family_repair_order_ablation} shows that
reversing the equation-repair and projection stages can reintroduce previously
resolved violations.

\paragraph{Logical-dependency enforcement.}
For each violated LD, we keep the determinant fixed and replace only the
dependent value with a value from the admissible set specified by its matching
value-table entry. We assign singleton sets deterministically and otherwise
sample from the reference conditional distribution restricted to the set.
This satisfies the rule where a repair is defined and favors
reference-like conditional frequencies rather than guaranteeing marginal
preservation.
For overlapping rules, we intersect admissible sets per dependent column set
and topologically process the acyclic determinant-to-dependent graph. Omitted
determinant configurations remain outside the rule's scope.

\paragraph{Equational-constraint enforcement.}
A checker identifies violations but does not specify a repair. For each
equation $c$ and possible target $j$, an LLM agent generates
$g_{c\rightarrow j}$ (implemented as \texttt{fix(df)}) to reconstruct $j$ from
the other participating columns. We validate it on the complete reference
table using the protocol in Section~\ref{sec:constraint_discovery} and retain
it only if $\operatorname{VR}(c)\leq\tau_{\mathrm{vio}}$ under the frozen
checker; otherwise, it is revised or discarded.
For equations that share columns, the KS-complement-guided heuristic described
in Appendix~\ref{app:equational_repair} jointly selects the repair targets and
their execution order from all feasible complete schedules. It first maximizes
the worst predicted change in marginal KS complement, avoiding schedules that
severely degrade any single column, and then selects the remaining schedule
with the largest total predicted change.
We use the KS complement because preserving marginal distributions 
is important for downstream statistical utility. Appendix~\ref{app:equational_repair_order_ablation}
evaluates this heuristic against randomized target and order selection
constrained to retain a complete repair schedule. After each repair, participating columns are protected and
resolved equations are revalidated.

\paragraph{Linear-inequality enforcement.}
We jointly project each violating row onto the common linear feasible
region. \citet{stoian2024how} sequentially correct features using a specified
variable order, whereas EVS uses order-free Euclidean projection
\citep{afonja2026from}. However, both P-DGM and EVS enforce linear inequalities
in isolation; neither method is able to coordinate the projection while preserving 
previously repaired
equational constraints. Our joint convex program instead minimizes
scale-normalized distortion while fixing columns in repaired equations,
yielding an order-independent global optimum that preserves those repairs.
Let $\widetilde{\mathbf{x}}_i$ be the row after the preceding stages,
$S_{\mathrm{num}}$ its numerical-column indices, and
$\mathbf{z}=(z_j)_{j\in S_{\mathrm{num}}}\in
\mathbb{R}^{|S_{\mathrm{num}}|}$. Stack the canonicalized inequalities as
$A\mathbf{z}\leq\mathbf{b}$. Let
$P\subseteq S_{\mathrm{num}}$ contain the equation-participating and
zero-variance columns, and let $U=S_{\mathrm{num}}\setminus P$:
\[
    \begin{aligned}
    \widehat{\mathbf{x}}_{i,S_{\mathrm{num}}}
    \in
    \underset{\mathbf{z}\in\mathbb{R}^{|S_{\mathrm{num}}|}}{\arg\min}
    \quad&
    \sum_{j\in U}
    \left(\frac{z_j-\widetilde{x}_{ij}}{s_j}\right)^2 \\
    \text{subject to}\quad&
    A\mathbf{z}\leq\mathbf{b}, \\
    &
    z_j=\widetilde{x}_{ij}
    \quad \forall j\in P .
    \end{aligned}
\]
For $j\in U$, $s_j>0$ is the reference standard deviation. Integer-valued
columns use their continuous relaxation; this stage does not enforce
integrality.  

Finally, after three stages of repairing, all family-specific validators recheck the
postprocessed table to ensure the constraints are all satisfied.

%% file: experiments.tex
\section{Evaluation}

\subsection{\texorpdfstring{Executable Constraint Discovery}{Executable Constraint Discovery}}
\label{sec:constraint_discovery_evaluation}

We evaluate executable discovery on three curated public datasets, each representing a different constraint family: \dataset{NBA} contains 12 equations relating derived basketball statistics; \dataset{URL} contains 17 inequalities over numerical webpage features; and a categorical version of \dataset{Anxiety} contains 13 logical dependencies (LDs) encoding implications and admissible mappings.
Using separate benchmarks provides
family-specific curated constraints and controlled audit tasks for each typed
representation; Section~\ref{sec:end_to_end_evaluation} evaluates the full
pipeline when multiple families are discovered and enforced together. 
For five 70/30 discovery--audit splits, both methods see only
the discovery partition; hidden audit rows form balanced contrastive tasks for
every constraint.

We access GPT-5.6 Luna (\citealp{openai2026gpt56luna}) and Claude Sonnet 5
(\citealp{anthropic2026claudesonnet5}) through their APIs and use each as the LLM
backbone for both methods. The baseline makes one direct-prompt call without
verification tools or revision; ours runs the full tool-grounded agentic
workflow from Section~\ref{sec:constraint_discovery}. 
We report held-out
violation-detection accuracy, precision, and recall 
using a support-weighted macro-average across constraint-specific tasks.
Appendix~\ref{app:constraint_discovery_evaluation} summarizes the annotations
and task construction.

\input{constraint_discovery_performance}

\textbf{Our complete executable validation-and-revision workflow achieves higher mean held-out violation-detection scores in all 18 family--backbone--metric comparisons.}
Table~\ref{tab:constraint-discovery-performance} shows the largest gains for LDs: precision and recall each rise from .442 to .974 with GPT-5.6 and .963 with Claude-5. Linear discovery also improves substantially, especially with Claude-5, whose accuracy increases from .815 to .959 and recall from .633 to .919. Even against near-ceiling equational baselines, our workflow reaches 1.000 on all three metrics with GPT-5.6 and .998 accuracy, 1.000 precision, and .996 recall with Claude-5. The simultaneous precision and recall gains indicate better discrimination between valid audit rows and contrastive violations; gains across both backbones suggest that the benefit is not model-specific.

\subsection{End-to-End Constraint-Aware Generation}
\label{sec:end_to_end_evaluation}

\paragraph{Experimental setup.}
We evaluate seven public datasets (Details in Appendix~\ref{app:datasets}, \dataset{Anxiety}
exluded since it's purely categorical and has no target column)
 using four generators that span distinct
modeling paradigms: Gaussian Copula (statistical)
\citep{patki2016synthetic}, CTGAN (adversarial) and TVAE (variational
autoencoding) \citep{xu2019modeling}, and TabDDPM (diffusion-based)
\citep{kotelnikov2023tabddpm}. For each of three independent 70/30 splits, we
train every generator once and draw three synthetic tables. Constraints are
independently discovered from each real training partition
using our agentic workflow with GPT-5.6 Luna backbone and default
hyperparameters and applied to the
corresponding synthetic outputs, yielding 252 paired raw--postprocessed evaluations.
Across datasets and splits, the system discovers 0--16 LD constraints, 0--12
equational constraints, and 0--29 linear inequalities per dataset;
Table~\ref{tab:end-to-end-results} reports the dataset-specific counts and
split-dependent ranges.
Appendix~\ref{app:end_to_end_details} provides the full protocol and
configuration details.

\paragraph{Metrics.}
CVR is the fraction of rows violating at least one constraint (perfect:
$\mathrm{CVR}=0$), whereas sCVC measures the density of violations across all
row--constraint pairs. Equational $R^2$ measures consistency with discovered
equations (perfect: $R^2=1$), and linear feasibility distance (LFD) measures
distance from the jointly feasible linear region (perfect: $\mathrm{LFD}=0$).
We assess downstream utility using the Train on Synthetic, Test on
Real (TSTR) protocol \citep{stoian2025survey}. Each dataset table contains a
target column: three define classification tasks, evaluated using ROC-AUC, and
four define regression tasks, evaluated using $R^2$. 
We assess univariate
marginal fidelity using Column Shapes.
Figure~\ref{fig:utility-fidelity-preservation} reports absolute raw and
postprocessed scores; definitions are in Appendix~\ref{app:end_to_end_details}.

\input{end_to_end_results}

\begin{figure}[t]
  \centering
  \includegraphics[width=\columnwidth]{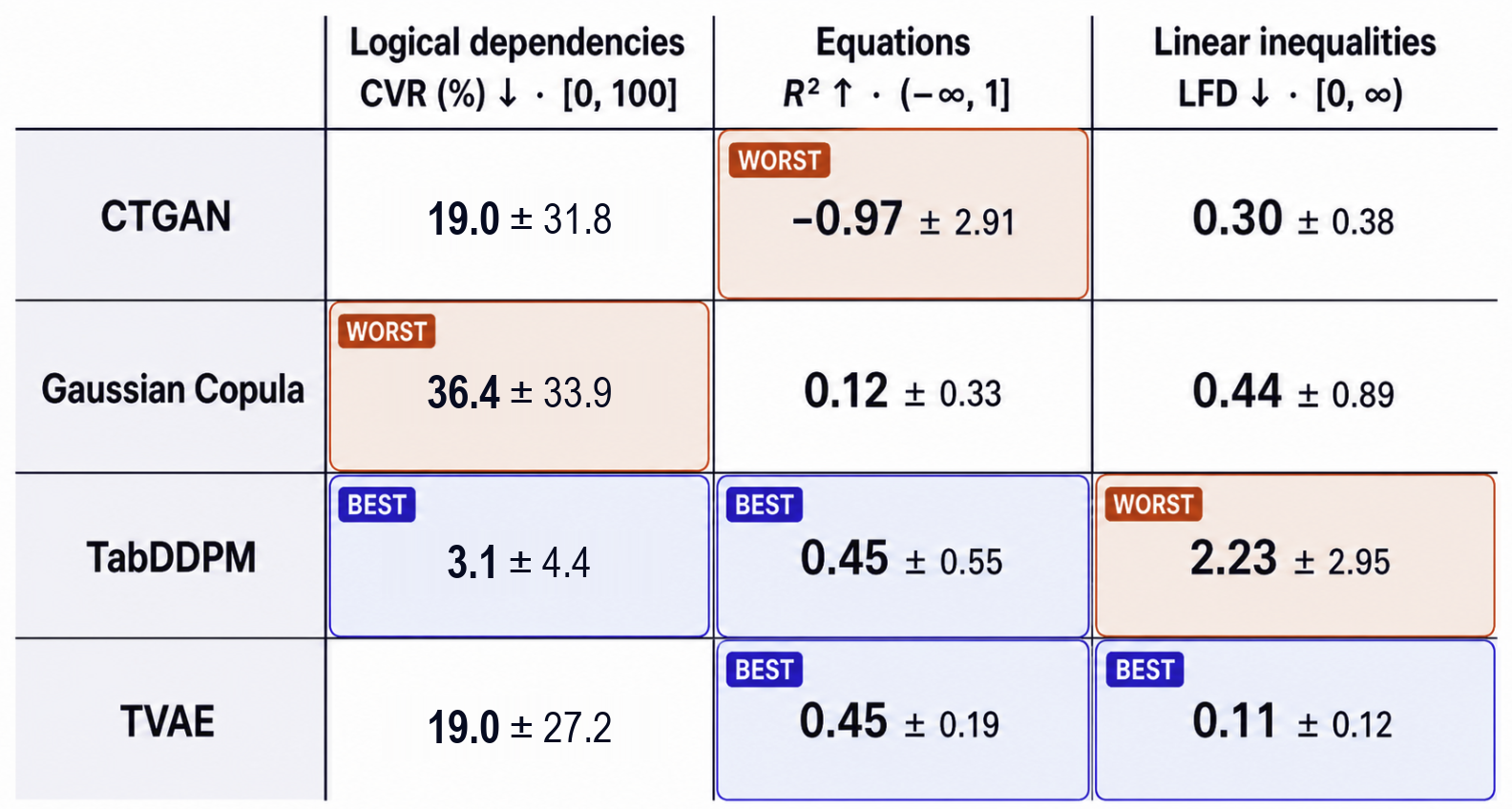}
  \caption{Generator-wise raw constraint metrics (mean $\pm$ standard
  deviation across applicable datasets). Blue and vermillion cells mark the
  best and worst means, respectively, within each constraint family.}
  \label{fig:raw-cvr-by-generator}
  \vspace{-0.1in}
\end{figure}

\textbf{Raw generators exhibit distinct, constraint-specific failure profiles.}
Figure~\ref{fig:raw-cvr-by-generator} reveals reversals in relative performance
across families. TabDDPM has the lowest LD CVR (3.1\%) but the highest LFD
(2.23), whereas TVAE has the lowest LFD (0.11); both attain the highest
equational $R^2$ (0.45) at the reported precision. Conversely, CTGAN is weakest
on equations ($R^2=-0.97$), and Gaussian Copula has the highest LD CVR (36.4\%).
Thus, preserving one family well does not predict preservation of another.
Generator selection alone is therefore insufficient for structural validity,
motivating family-aware enforcement that remains agnostic to generator
architecture.

\begin{figure}[!t]
  \centering
  \includegraphics[width=\columnwidth]{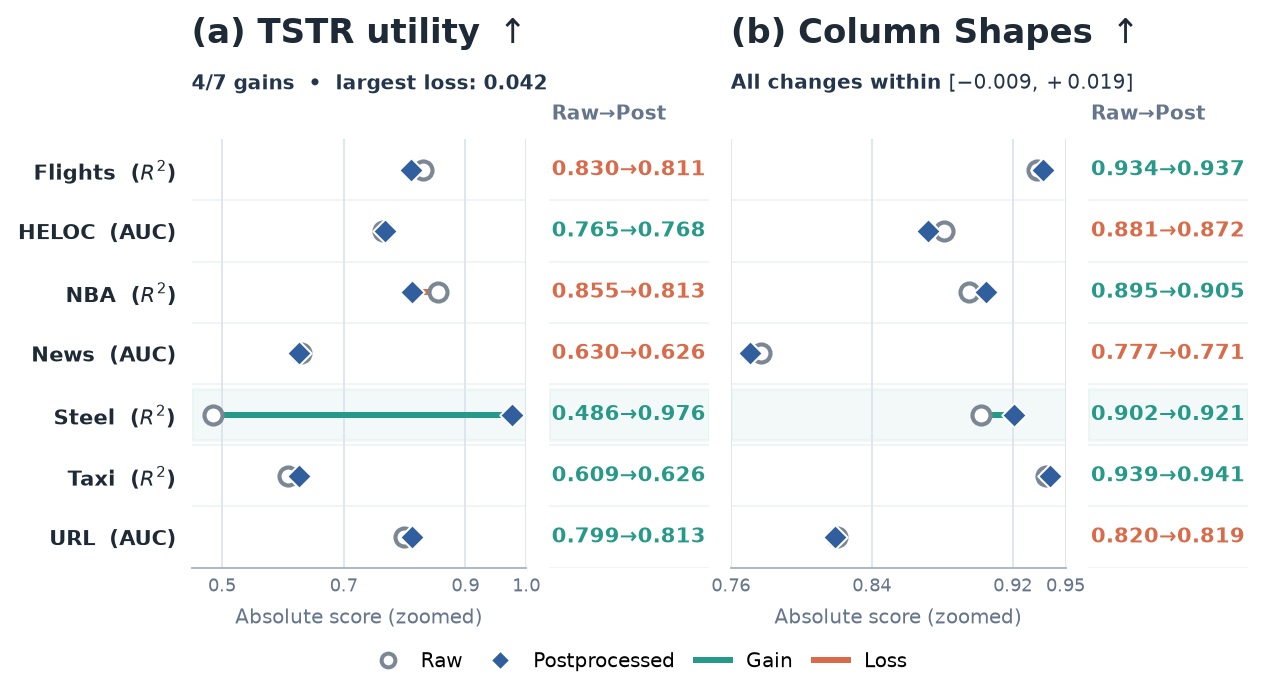}
  \caption{Raw-to-postprocessed TSTR utility and Column Shapes. TSTR uses
  ROC-AUC for classification datasets and $R^2$ otherwise; only within-dataset
  changes are comparable.}
  \label{fig:utility-fidelity-preservation}
  \vspace{-0.1in}
\end{figure}

\textbf{Postprocessing exactly satisfies every applicable family while
largely preserving utility and fidelity.} Table~\ref{tab:end-to-end-results}
shows substantial violations in raw outputs: equational CVR ranges from 98.2\%
to 100\%, LD CVR reaches 65.9\%, and linear CVR reaches 77.8\%. Postprocessing
nevertheless reaches the exact optimum for every applicable metric: CVR, sCVC,
and LFD fall to zero, while equational consistency rises to $R^2=1$. Reaching
these targets jointly across heterogeneous family combinations shows that
repair does not merely shift violations from one constraint type to another.
Figure~\ref{fig:utility-fidelity-preservation} grounds these changes in their
absolute baselines: mean utility improves on four of seven datasets
(\dataset{Steel}: $0.486\rightarrow0.976$) and decreases by at most $0.042$,
while Column Shapes changes by only $-0.009$ to $+0.019$. Thus, exact
enforcement largely preserves univariate marginal fidelity and can sometimes
benefit downstream prediction.

\subsection{Ablation Studies}
\label{sec:ablation_studies}

\begin{figure}[!t]
    \centering
    \includegraphics[width=\columnwidth]{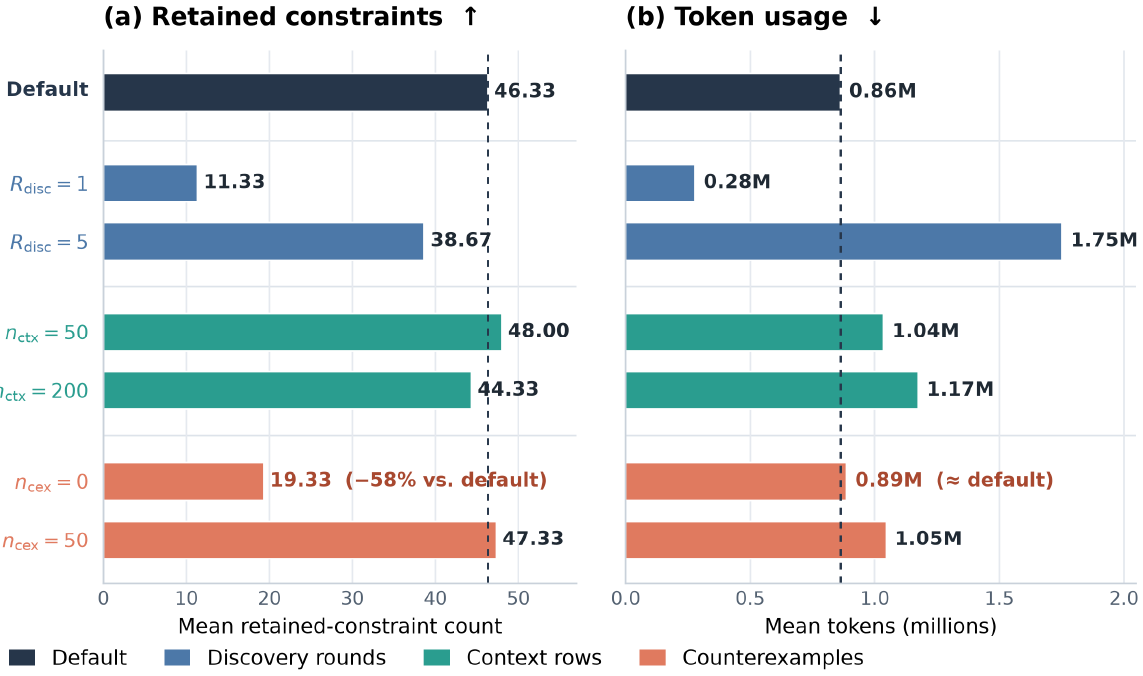}
    \caption{Discovery sensitivity on \dataset{News} (three-split means;
    GPT-5.6 Luna). Each variant changes one component of the default
    $(3,100,20)$; dashed lines mark it. Higher yield and lower token use are
    better.}
    \label{fig:discovery-hyperparameter-sensitivity}
\end{figure}

\paragraph{Discovery configuration.}
\textbf{Counterexamples and repeated discovery increase retained yield, but
larger budgets show diminishing returns.}
On \dataset{News} (Figure~\ref{fig:discovery-hyperparameter-sensitivity}),
removing counterexamples and using one rather than three rounds lower mean
validated-hypothesis yield from 46.33 to 19.33 and 11.33, respectively. Five
rounds more than double token use but yield only 38.67; alternative context and
larger counterexample budgets offer at most marginal gains. Thus, the default
$(3,100,20)$ is the best observed cost--yield trade-off among the tested
settings, not a universal optimum
(Appendix~\ref{app:discovery_hyperparameter_sensitivity}).

\paragraph{Equational repair scheduling.}
\textbf{At equal constraint satisfaction, KS-complement guidance has higher
mean marginal fidelity than random feasible scheduling.}
Both attain zero equational CVR on 180 matched inputs from five datasets. With
inputs, constraints, and repair functions fixed, KS guidance has higher mean
paired Column Shapes on four datasets by $0.020$--$0.053$, ties on
\dataset{News}, and yields an equal-weight mean gain of $0.025$
(Appendix~\ref{app:equational_repair_order_ablation}).

\paragraph{Cross-family repair order.}
\textbf{Projection last avoids observed cross-family interference.}
On \dataset{NBA}, with $E$ denoting equational repair and $L$ linear-inequality
projection, $E{\rightarrow}L$ yields zero final CVR for both families and zero
LFD in all 36 matched settings. The reverse leaves linear violations in 24
settings (66.7\%; CVR $4.574\%$; LFD $0.863$), while equational CVR remains
zero. This supports projection last when numerical families overlap
(Appendix~\ref{app:cross_family_repair_order_ablation}).

%% file: constraint_discovery_performance.tex
\begin{table}[t]
  \centering
  \scriptsize
  \setlength{\tabcolsep}{1.5pt}
  \renewcommand{\arraystretch}{1.0}
  \begin{tabular*}{\columnwidth}{@{\extracolsep{\fill}}lllccc@{}}
    \toprule
    Type & Model & Method & Acc. & Prec. & Rec. \\
    \midrule
    \multirow{4}{*}{Linear}
      & \multirow{2}{*}{GPT-5.6}
      & Baseline & .842 (.016) & .741 (.032) & .684 (.032) \\
      & & Ours  & \textbf{.917} (.096) & \textbf{.911} (.181)
              & \textbf{.833} (.193) \\
    \cmidrule(lr){2-6}
      & \multirow{2}{*}{Claude-5}
      & Baseline & .815 (.033) & .729 (.089) & .633 (.067) \\
      & & Ours  & \textbf{.959} (.016) & \textbf{.975} (.031)
              & \textbf{.919} (.032) \\
    \midrule
    \multirow{4}{*}{Equational}
      & \multirow{2}{*}{GPT-5.6}
      & Baseline & .975 (.037) & .950 (.075) & .950 (.075) \\
      & & Ours  & \textbf{1.000} (.000) & \textbf{1.000} (.000)
              & \textbf{1.000} (.000) \\
    \cmidrule(lr){2-6}
      & \multirow{2}{*}{Claude-5}
      & Baseline & .960 (.025) & .933 (.037) & .919 (.051) \\
      & & Ours  & \textbf{.998} (.002) & \textbf{1.000} (.000)
              & \textbf{.996} (.003) \\
    \midrule
    \multirow{4}{*}{LD}
      & \multirow{2}{*}{GPT-5.6}
      & Baseline & .721 (.000) & .442 (.000) & .442 (.000) \\
      & & Ours  & \textbf{.987} (.016) & \textbf{.974} (.032)
              & \textbf{.974} (.032) \\
    \cmidrule(lr){2-6}
      & \multirow{2}{*}{Claude-5}
      & Baseline & .721 (.000) & .442 (.000) & .442 (.000) \\
      & & Ours & \textbf{.982} (.007) & \textbf{.963} (.014)
              & \textbf{.963} (.014) \\
    \bottomrule
  \end{tabular*}
  \caption{Held-out behavioral violation-detection results over five discovery--audit splits (mean and standard deviation). The audit tasks are constructed from 17 inequalities, 12 equations, and 13 logical dependencies.}
  \label{tab:constraint-discovery-performance}
\end{table}

%% file: end_to_end_results.tex
\begin{table*}[t]
  \centering
  \scriptsize
  \setlength{\tabcolsep}{3.2pt}
  \renewcommand{\arraystretch}{1.20}
  \begin{tabular}{@{}l@{\hspace{0.6em}}lccccccc@{}}
    \toprule
    \multicolumn{2}{l}{Metric} & \dataset{Flights} & \dataset{HELOC} & \dataset{NBA} &
    \dataset{News} & \dataset{Steel} & \dataset{Taxi} & \dataset{URL} \\
    \midrule
    \multicolumn{2}{l}{Constraint \# (LD/Eq/Lin)}
      & 3/7/2--5 & 1/0/11--13 & 0/10--12/3--11 & 7--16/4/0--17
      & 1/2/11--25 & 2/2/0 & 2/0/17--29 \\
    \midrule
    \multirow{2}{*}{LD} & CVR (\%) $\downarrow$
      & 65.9 $\rightarrow$ \textbf{0} & 4.0 $\rightarrow$ \textbf{0} & --
      & 20.2 $\rightarrow$ \textbf{0} & 11.7 $\rightarrow$ \textbf{0}
      & 12.2 $\rightarrow$ \textbf{0} & 2.2 $\rightarrow$ \textbf{0} \\
      & sCVC (\%) $\downarrow$
      & 40.0 $\rightarrow$ \textbf{0} & 4.0 $\rightarrow$ \textbf{0} & --
      & 2.2 $\rightarrow$ \textbf{0} & 11.7 $\rightarrow$ \textbf{0}
      & 6.3 $\rightarrow$ \textbf{0} & 1.1 $\rightarrow$ \textbf{0} \\
    \addlinespace[2pt]
    \multirow{3}{*}{Eq.} & CVR (\%) $\downarrow$
      & 100.0 $\rightarrow$ \textbf{0} & -- & 99.9 $\rightarrow$ \textbf{0}
      & 100.0 $\rightarrow$ \textbf{0} & 98.2 $\rightarrow$ \textbf{0}
      & 100.0 $\rightarrow$ \textbf{0} & -- \\
      & sCVC (\%) $\downarrow$
      & 99.3 $\rightarrow$ \textbf{0} & -- & 94.4 $\rightarrow$ \textbf{0}
      & 86.1 $\rightarrow$ \textbf{0} & 63.7 $\rightarrow$ \textbf{0}
      & 96.6 $\rightarrow$ \textbf{0} & -- \\
      & $R^2$ $\uparrow$
      & 0.31 $\rightarrow$ \textbf{1} & -- & $-1.30$ $\rightarrow$ \textbf{1}
      & $-0.19$ $\rightarrow$ \textbf{1} & 0.65 $\rightarrow$ \textbf{1}
      & 0.58 $\rightarrow$ \textbf{1} & -- \\
    \addlinespace[2pt]
    \multirow{3}{*}{Lin.} & CVR (\%) $\downarrow$
      & 0.0 $\rightarrow$ \textbf{0} & 58.7 $\rightarrow$ \textbf{0}
      & 8.0 $\rightarrow$ \textbf{0} & 42.3 $\rightarrow$ \textbf{0}
      & 5.7 $\rightarrow$ \textbf{0} & -- & 77.8 $\rightarrow$ \textbf{0} \\
      & sCVC (\%) $\downarrow$
      & 0.0 $\rightarrow$ \textbf{0} & 7.9 $\rightarrow$ \textbf{0}
      & 0.7 $\rightarrow$ \textbf{0} & 5.7 $\rightarrow$ \textbf{0}
      & 0.2 $\rightarrow$ \textbf{0} & -- & 11.1 $\rightarrow$ \textbf{0} \\
      & LFD $\downarrow$
      & 0.000 $\rightarrow$ \textbf{0} & 1.169 $\rightarrow$ \textbf{0}
      & 0.029 $\rightarrow$ \textbf{0} & 1.483 $\rightarrow$ \textbf{0}
      & 0.020 $\rightarrow$ \textbf{0} & -- & 1.912 $\rightarrow$ \textbf{0} \\
    \bottomrule
  \end{tabular}
  \caption{End-to-end constraint enforcement across seven public datasets.
  Constraint-metric cells show raw $\rightarrow$ postprocessed means; exact
  optima are written as $0$ or $1$. Dataset means average samples within each
  generator and split, then generators and splits equally. Constraint counts
  span splits; dashes denote unavailable families.}
  \label{tab:end-to-end-results}
\end{table*}

%% file: conclusion.tex
\section{Conclusion}

We presented a unified framework that turns LLM-proposed equations, linear inequalities, and logical dependencies into executable hypotheses and coordinates their post-hoc enforcement on outputs from unchanged generators. Across multiple LLMs, datasets, and generator families, the complete validation-and-revision workflow improves held-out violation detection over direct prompting, while postprocessing yields zero measured violations under retained, applicable validators with generally preserved utility and univariate marginal fidelity. Coupling counterexample-grounded discovery with coordinated repair offers a practical path to structurally reliable tabular synthesis.

%% file: limitations.tex
\section*{Limitations}

Our method is limited to inter-column constraints evaluated within each
record. It does not model inter-row constraints, such as uniqueness across
records, or cell-level validity and formatting rules, such as regular-expression
patterns and domain-specific semantic types supported by SDV
\citep{datacebo2026sdv}. Extending discovery and enforcement to these
constraint families is outside the scope of this work.

Full-table validation establishes empirical consistency, not semantic
correctness or complete discovery. Zero measured violations therefore applies
only to retained, applicable validators and excludes uncovered LD
configurations or unavailable repairs.

In all experiments, we supply rich metadata, including dataset- and
column-level descriptions, which the LLM discovery agents use to propose
constraint hypotheses. In practice, however, many tables may lack such
annotations or have ambiguous and noisy column semantics, potentially reducing
discovery accuracy. Evaluating robustness to limited-quality metadata remains
future work.

%% file: ethics_statement.tex
\section*{Ethics Statement}

We use public tabular datasets. Some datasets nevertheless
represent potentially sensitive domains, including credit risk and mental
health. We use these datasets only for methodological evaluation and do not
attempt to identify individuals or support decisions about them.

The LLM components receive dataset metadata, column profiles, sampled records,
and verifier-selected counterexamples. Consequently, applying the
framework to confidential or personally identifiable data would require
appropriate authorization, data-protection measures, and careful consideration
before transmitting records to third-party model providers.

Structural validity does not guarantee privacy or fairness: discovered
constraints may encode historical biases and require domain
review. Generated Python must be securely sandboxed.

%% file: appendices.tex
\vspace*{1pt}

\input{constraint_hypothesis_representations}
\input{equational_repair_algorithm}
\input{datasets}
\input{constraint_discovery_details}
\input{end_to_end_constraint_aware_generation}
\input{ablation_studies}

%% file: constraint_hypothesis_representations.tex
\section{Constraint Discovery and Enforcement Details}
\label{app:constraint_discovery_enforcement}

\subsection{Constraint Hypothesis Representations}
\label{app:constraint_representations}

The discovery agent emits a different machine-executable representation for
each constraint family. The examples below show the concrete schemas referenced
in Section~\ref{sec:constraint_discovery}. We use a display-oriented rendering
for readability; in the equational example, the escaped line breaks stored in
the \texttt{check\_code} string are expanded into an indented Python block.

\refstepcounter{constraintexample}
\label{ex:eq-hypothesis}
\begin{tcblisting}{
    constraint card,
    colback=eqbackground,
    colframe=eqaccent!75!black,
    colbacktitle=eqaccent!13,
    borderline west={2.2pt}{0pt}{eqaccent},
    title={Example~\theconstraintexample: Equational},
    listing options={style=constraintpython}
}
id:
  "eq_absolute_title_subjectivity"

description:
  "abs_title_subjectivity is the absolute
   distance of title_subjectivity from 0.5."

columns:
  ["abs_title_subjectivity", "title_subjectivity"]

check_code:
def check(df):
    absolute = pd.to_numeric(
        df["abs_title_subjectivity"],
        errors="coerce",
    )
    subjectivity = pd.to_numeric(
        df["title_subjectivity"],
        errors="coerce",
    )
    finite = (
        absolute.notna()
        & subjectivity.notna()
        & np.isfinite(absolute)
        & np.isfinite(subjectivity)
    )
    error = absolute - (subjectivity - 0.5).abs()
    return finite & error.abs().le(1e-9)
\end{tcblisting}

\refstepcounter{constraintexample}
\label{ex:linear-hypothesis}
\begin{tcblisting}{
  constraint card,
  colback=linbackground,
  colframe=linaccent!75!black,
  colbacktitle=linaccent!13,
  borderline west={2.2pt}{0pt}{linaccent},
  title={Example~\theconstraintexample: Linear inequality}
}
{
"id": "avg_negative_polarity_ge_min",
"description": "Average negative-word polarity is at least the minimum negative-word polarity.",
"columns": [
  "avg_negative_polarity",
  "min_negative_polarity"
],
"coefficients": {
  "avg_negative_polarity": 1,
  "min_negative_polarity": -1
},
"sense": ">=",
"rhs": 0
}
\end{tcblisting}

\refstepcounter{constraintexample}
\label{ex:logical-value-table}
\begin{tcblisting}{
    constraint card,
    colback=catbackground,
    colframe=cataccent!75!black,
    colbacktitle=cataccent!13,
    borderline west={2.2pt}{0pt}{cataccent},
    title={Example~\theconstraintexample: Logical dependency}
}
{
  "id": "plan_severity_support_channel",
  "description": "The subscription plan and issue severity jointly restrict the available support channels.",
  "determinants": [
    "subscription_plan",
    "issue_severity"
  ],
  "dependent": "support_channel",
  "value_table": [
    {
      "determinant_values": [
        ["free", "standard"],  # IF plan: free OR standard
        ["low"]                #  AND severity: low
      ],
      "dependent_values": [    # THEN channel is one of
        "community_forum",
        "email"
      ]
    },
    {
      "determinant_values": [
        ["premium"],        # IF plan: premium
        ["low", "medium"]   #  AND severity: low OR medium
      ],
      "dependent_values": [ # Then channel is one of
        "email",
        "live_chat"
      ]
    },
    {
      "determinant_values": [
        ["premium", "enterprise"], # IF plan: premium OR enterprise
        ["high", "critical"]       #  AND severity: high OR critical
      ],
      "dependent_values": [        # Then channel is one of
        "priority_email",
        "live_chat",
        "phone"
      ]
    }
  ]
}
\end{tcblisting}

\refstepcounter{constraintexample}
\label{ex:fd-shorthand}
\begin{tcblisting}{
    constraint card,
    colback=catbackground,
    colframe=cataccent!75!black,
    colbacktitle=cataccent!13,
    borderline west={2.2pt}{0pt}{cataccent},
    title={Example~\theconstraintexample: Logical dependency}
}
{
  "id": "cat_fd_001",
  "description": "Each airport code determines its city.",
  "determinants": ["airport_code"],
  "dependent": "city",
  "value_table": [
    {
      "determinant_values": [["JFK"]], # IF airport_code: JFK
      "dependent_values": ["New York"] # THEN city: New York
    }, 
    {
      "determinant_values": [["SFO"]], 
      "dependent_values": ["San Francisco"] 
    },
    {
      "determinant_values": [["SEA"]],
      "dependent_values": ["Seattle"]
    },
    ...
  ]
}
\end{tcblisting}

\clearpage
\onecolumn
\subsection{Discovery LLM Prompts}
\label{app:discovery_prompts}

We use a family-specific prompt for each constraint representation. The
templates below show the instructions supplied to the discovery agent; braces
denote values inserted by the host at run time.

\begin{tcblisting}{
    constraint card,
    colback=eqbackground,
    colframe=eqaccent!75!black,
    colbacktitle=eqaccent!13,
    borderline west={2.2pt}{0pt}{eqaccent},
    title={Equational constraint discovery prompt},
    listing options={style=constraintprompt}
}
[System]
You are a meticulous tabular-data semantics auditor. Identify domain-defensible, deterministic numerical equations between columns using the dataset description, column metadata, units, representative records, and full-data verification. Prefer correctness, simplicity, and generality over the number of constraints.

A valid constraint expresses one row-wise numerical relationship that should hold for essentially every row.
Exclude correlations, trends, inequalities, distributional or conditional rules, and formulas fitted to the sample.

For each candidate, provide exactly:
1. id: unique snake_case identifier;
2. description: concise statement of the equation;
3. columns: all and only referenced numerical columns; and
4. check_code: vectorized check(df) returning an index-aligned Boolean pandas Series, with True for satisfying rows.

The checker must not import, mutate df, perform I/O, access external state, use row indices, hard-code examples or exceptions, fit parameters, or use full-dataset aggregates.
Use exact comparisons for integer identities and only a small, semantically justified tolerance for floating-point identities.

[User]
Dataset description:
{dataset_description}

Column profiles:
{numerical_column_profiles}

Data sample:
{sampled_rows}

Already accepted hypotheses (do not repeat or rearrange):
{accepted_hypotheses_json}

Previously rejected hypotheses (do not repeat or revise):
{rejected_hypotheses_json}

Discover as many non-speculative constraints as the evidence supports and submit them together in one verifier call.
\end{tcblisting}

\clearpage
\begin{tcblisting}{
    constraint card,
    colback=catbackground,
    colframe=cataccent!75!black,
    colbacktitle=cataccent!13,
    borderline west={2.2pt}{0pt}{cataccent},
    title={Logical dependency discovery prompt},
    listing options={style=constraintprompt}
}
[System]
You are a meticulous tabular-data semantics auditor. Discover semantically meaningful logical dependencies among categorical columns using column metadata, representative records, and full-data evidence.

A constraint has one or more determinant columns, exactly one dependent column, and a value table. Each table entry lists admissible values for every determinant and for the dependent. A row is applicable when its determinant values match an entry and violates the constraint when its dependent value is not admissible; unmatched rows do not violate it.

For a large exact dependency, you may submit an empty value table; the host then builds the complete mapping using the majority dependent value for each observed determinant configuration. Conditional and multi-admissible mappings require an explicit value table.

For each candidate, provide exactly:
1. id: unique snake_case identifier;
2. description: concise statement of the dependency;
3. columns: all and only referenced categorical columns; and
4. constraint_representation: {logical_dependency_DSL_format}

Prioritize semantically grounded relationships such as codes determining names, geographic identifiers determining regions, classifications determining labels, and stable status lookups. 
Prefer minimal determinant sets. Avoid row keys, identifiers, nearly unique determinants, accidental sample patterns, tautological tables, and arbitrary groupings created only to reduce violations.

[User]
Dataset description:
{dataset_description}

Column profiles:
{categorical_column_profiles}

Data sample:
{sampled_rows}

Already accepted hypotheses (do not repeat or rearrange):
{accepted_hypotheses_json}

Previously rejected hypotheses (do not repeat or revise):
{rejected_hypotheses_json}

Inspect and analyze promising logical dependencies. Finish with the required structured completion when no additional semantically defensible logical dependency remains.
\end{tcblisting}

\clearpage
\begin{tcblisting}{
    constraint card,
    colback=linbackground,
    colframe=linaccent!75!black,
    colbacktitle=linaccent!13,
    borderline west={2.2pt}{0pt}{linaccent},
    title={Linear-inequality discovery prompt},
    listing options={style=constraintprompt}
}
[System]
You are a meticulous tabular-data semantics auditor. Discover semantically meaningful, row-wise linear inequalities among numerical columns.

A useful constraint is a universal semantic relationship expected to hold for essentially every row. Prioritize simple inequalities justified by column definitions, especially totals covering subsets, minimum--average--maximum orderings, lengths or capacities covering components, and other whole--part relationships.

For each candidate, provide exactly:
1. id: unique snake_case identifier;
2. description: concise statement of the linear inequality;
3. columns: all and only referenced numerical columns; and
4. constraint_representation: {linear_DSL_format}

Use small integer coefficients when possible. Exclude correlations, trends, fitted regressions, quantile or distributional claims, conditional rules, unlisted columns, and arbitrary constants selected from sample extrema.  Do not submit positive scalar multiples of an existing inequality, or repeat, weaken, or rescale an accepted or rejected hypothesis.

[User]

Dataset description:
{dataset_description}

Column profiles:
{numerical_column_profiles}

Data sample:
{sampled_rows}

Already accepted hypotheses (do not repeat or rearrange):
{accepted_hypotheses_json}

Previously rejected hypotheses (do not repeat or revise):
{rejected_hypotheses_json}

Discover all distinct, semantically defensible linear inequalities supported by the evidence. Avoid previous hypotheses and scaled rewrites.
\end{tcblisting}

\clearpage
\begin{tcblisting}{
    constraint card,
    colback=eqbackground,
    colframe=eqaccent!75!black,
    colbacktitle=eqaccent!13,
    borderline west={2.2pt}{0pt}{eqaccent},
    title={Constraint refinement prompt},
    listing options={style=constraintprompt}
}
[User]
Refinement round {refinement_round} of {max_refinement_rounds} for discovery phase {phase}.

Column profiles:
{involved_column_profiles}

Samples that failed at validation during the previous round:
{counterfactual_samples}

Candidate and full verification history, including violating samples from every previous attempt:
{candidate_history_json}

Submit exactly one revision with the same constraint ID. If no principled revision remains, finish with `rejected_hypotheses` listing that ID and a concrete reason.
\end{tcblisting}
\twocolumn

%% file: equational_repair_algorithm.tex
\subsection{KS-Guided Equational Repair}
\label{app:equational_repair}
Given a synthetic table $S$, a real training table $R$, and a set of
equational constraints $\mathcal{C}$, we need to choose which column to repair
for each constraint and the order in which to perform the repairs. Different
choices may satisfy the same constraints but have different effects on the
quality of the synthetic data. We measure the similarity between a real column
$R_j$ and a synthetic column $S_j$ using the KS complement:
\[
    Q(R_j,S_j)=1-D_{\mathrm{KS}}(R_j,S_j),
\]
where a larger value indicates more similar marginal distributions.

For each constraint $c$, let $\mathcal{T}_c$ be the set of columns that can be
repaired. For each possible target $j\in\mathcal{T}_c$, we apply the repair
$f_{c,j}$ to the original synthetic table and compute
\[
    \Delta_{c,j}
    =
    Q\!\left(R_j,f_{c,j}(S)_j\right)-Q(R_j,S_j).
\]
The value $\Delta_{c,j}$ estimates how repairing column $j$ changes its
marginal similarity to the real data. A positive value indicates an expected
improvement, while a negative value indicates an expected loss.

Repair order matters because different constraints may share columns. Once a
constraint is repaired, we freeze all columns involved in that constraint.
Later repairs may use these columns as inputs but cannot overwrite them. This
ensures that a later repair does not invalidate a constraint that has already
been satisfied.

A repair schedule can be viewed as a path containing one target choice for
each constraint. Among all valid paths, we choose the one whose lowest
$\Delta_{c,j}$ is as high as possible:
\[
    \tau^\star
    =
    \max_{P\in\mathcal{P}}
    \min_{(c,j)\in P}\Delta_{c,j},
\]
where $\mathcal{P}$ is the set of valid complete repair schedules. In other
words, we seek the path with the best worst repair. This avoids choosing a
schedule that contains one highly damaging repair, even if its other repairs
have large improvements.

If several paths have the same best minimum value, we choose the one with the
largest total score:
\[
    P^\star
    \in
    \arg\max_{\substack{P\in\mathcal{P}:\\
    \min_{(c,j)\in P}\Delta_{c,j}\geq\tau^\star}}
    \sum_{(c,j)\in P}\Delta_{c,j}.
\]
The first objective protects against a poor individual repair, while the
second selects the path with the best overall predicted effect.

We solve this optimization problem by considering the distinct candidate
scores $\Delta_{c,j}$ in descending order. For each score threshold $\tau$, we
use dynamic programming to search for a complete dependency-safe schedule
using only repairs with $\Delta_{c,j}\geq\tau$. Each state records the set of
constraints already scheduled, which determines the columns that are frozen
and the repairs that remain available. The first threshold for which a
complete schedule exists gives the optimal minimum score $\tau^\star$. Among
all complete schedules satisfying this threshold, the dynamic program returns
the one with the largest sum of candidate scores. The selected repairs are
then applied sequentially in the resulting order.



%% file: datasets.tex
\section{Datasets}
\label{app:datasets}
For the eight public tabular datasets used in our experiments, we document
their brief descriptions, sources, and tasks in downstream utility evaluations.
Table~\ref{tab:dataset-summary} summarizes their Total-Train-Test sizes
and categorical and numerical column counts.

\begin{table}[H]
  \centering
  \small
  \setlength{\tabcolsep}{3.5pt}
  \renewcommand{\arraystretch}{1.08}
  \begin{tabular}{@{}lrrrrr@{}}
    \toprule
    Dataset & Total & Train & Test & Cat. & Num. \\
    \midrule
    \dataset{Flights} & 60,000 & 42,000 & 18,000 & 9 & 14 \\
    \dataset{HELOC}   &  9,084 &  6,358 &  2,726 & 3 & 21 \\
    \dataset{NBA}     & 14,062 &  9,843 &  4,219 & 0 & 26 \\
    \dataset{News}    & 38,458 & 26,920 & 11,538 & 9 & 45 \\
    \dataset{Steel}   & 34,993 & 24,495 & 10,498 & 3 &  7 \\
    \dataset{Taxi}    & 59,996 & 41,997 & 17,999 & 7 & 15 \\
    \dataset{URL}     & 11,430 &  8,001 &  3,429 & 50 & 37 \\
    \dataset{Anxiety} & 11,000 &  -     &  -     & 20 & 0 \\
    \bottomrule
  \end{tabular}
  \caption{Dataset sizes and column-type counts for the frozen end-to-end
  evaluation splits. Column counts include the utility target. Anxiety is not included in 
  the end-to-end evaluation task.}
  \label{tab:dataset-summary}
\end{table}

\paragraph{Flights.}
We use a uniformly selected 60,000-row processed sample of completed U.S.
flights from 2019--2023, sourced from the Kaggle Flight Delay and Cancellation
dataset.\footnote{\url{https://www.kaggle.com/datasets/patrickzel/flight-delay-and-cancellation-dataset-2019-2023}}
We remove cancelled or diverted flights; retain only rows whose origin and
destination airports and cities are among the 100 most frequent values; and
remove rows with missing or invalid clock-time or time-zone values. We then
convert local flight times to UTC minute-of-day values.
Its columns describe airlines, airports, route distance, scheduled and actual
UTC clock times, taxi and air durations, and delays. The utility task is
regression on \texttt{ARR\_DELAY}.

\paragraph{HELOC.}
The anonymized FICO Home Equity Line of Credit dataset is obtained through
Hugging Face.\footnote{\url{https://huggingface.co/datasets/mstz/heloc}}
Each row represents a homeowner's credit application, with 23 credit-bureau
attributes used to predict repayment performance over two years. The utility
task is binary classification of \texttt{is\_at\_risk}: 1 denotes ``Bad''
performance and 0 denotes ``Good.'' During preprocessing, we remove 588 rows
in which all 23 predictors equal the missing-value code $-9$. We also remove
787 internally inconsistent rows in which \texttt{nr\_total\_trades} is smaller
than at least one of its four component counts:
\texttt{number\_of\_satisfactory\_trades},
\texttt{nr\_trades\_initiated\_in\_last\_year},
\texttt{nr\_revolving\_trades\_with\_balance}, or 
\texttt{nr\_installment\_trades\_with\_balance}.

\paragraph{NBA.}
The NBA data come from the pbpstats totals API.\footnote{\url{https://api.pbpstats.com/docs}}
They contain player-season scoring statistics for 30 regular seasons, from
1996--97 through 2025--26. Counts are normalized per 100 offensive possessions
and describe shooting volume, efficiency, assisted scoring, putbacks, blocked
attempts, and related measures. We replace missing numerical values---mostly
sparse zero-count statistics---with zero. The utility task is regression on
\texttt{Usage}, the percentage of team possessions ending in the player's shot,
free-throw trip, or turnover.

\paragraph{News.}
The UCI Online News Popularity dataset contains
feature summaries for Mashable articles published from 2013--2015.\footnote{\url{https://archive.ics.uci.edu/dataset/332/online+news+popularity}}
Features cover article structure, keywords, channels, topics, links, and
sentiment. We remove 1,186 rows in which both the positive- and negative-word
rates are zero. We create the binary target \texttt{is\_popular}, assigning 1
to articles with at least 1,400 shares and 0 otherwise, and then remove
\texttt{shares} and the article URL. The utility task is binary classification
of \texttt{is\_popular}.

\paragraph{Steel.}
The Steel Industry Energy Consumption dataset records electricity use at
15-minute intervals over one calendar year.\footnote{\url{https://www.kaggle.com/datasets/csafrit2/steel-industry-energy-consumption}}
Its variables cover reactive power, power factors, CO$_2$ emissions, time of
day, weekday/weekend status, and operational load type. We remove 47 anomalous
rows from the continuous interval on January 2, 2018, from 08:15 to 20:30. The
interval contains 50 records; in the 47 removed records, \texttt{Usage\_kWh}
implies an expected CO$_2$ value between 0.01 and 0.07, whereas the recorded
CO$_2$ value remains zero. The utility task is regression on
\texttt{Usage\_kWh}, the active electricity consumption.

\paragraph{Taxi.}
The taxi data are a processed sample of 2015 NYC Green Taxi trips from the NYC
Taxi and Limousine Commission.\footnote{\url{https://www.nyc.gov/site/tlc/about/tlc-trip-record-data.page}}
We remove the row identifier, \texttt{VendorID}, the original pickup and
dropoff timestamps, and \texttt{Ehail\_fee}, which is entirely missing.
Columns describe locations, distance, passenger and fare codes, clock times,
duration, and itemized charges. The utility task is regression on
\texttt{total\_amount}, which is the sum of fare, tax, tip, toll,
extra, and improvement-surcharge of a trip.

\paragraph{URL.}
The Web Page Phishing Detection dataset is an exactly balanced benchmark with
5,715 legitimate and 5,715 phishing pages.\footnote{\url{https://doi.org/10.17632/c2gw7fy2j4.2}}
It combines URL syntax, fetched-page HTML content, and external-service features
collected in May 2020. We remove the source \texttt{url} field because it is
nearly unique, as well as \texttt{submit\_email}. The utility task is binary
classification of \texttt{status} as \texttt{legitimate} or \texttt{phishing}.

\paragraph{Anxiety.}
The Social Anxiety Dataset is an 11,000-row survey of anxiety severity and its
lifestyle, physiological, and clinical
correlates.\footnote{\url{https://www.kaggle.com/datasets/natezhang123/social-anxiety-dataset}}
Columns describe demographics, lifestyle habits, family history and recent
life events, physiological readings, treatment use, and self-reported stress
and anxiety scores. It is used for constraint discovery only, as our
logical-dependency benchmark, and has no utility task. Because logical
dependencies are categorical, we discretize its twelve numerical columns into
two to four ordinal bands each and derive \texttt{Occupation Group} from
\texttt{Occupation}, giving a view of 11,000 rows and 20 categorical columns.

%% file: constraint_discovery_details.tex
\section{Constraint Discovery Details}
\label{app:constraint_discovery_evaluation}

\subsection{Expert Annotation and Ground-Truth Constraints}
\label{app:expert_annotation}

We construct the ground truth through a two-annotator protocol. The first
annotator proposes candidate constraints from the dataset documentation,
schema, feature definitions, and domain semantics. The second annotator
independently reviews every candidate along three dimensions:
(i)~\emph{soundness}, whether the relation is satisfied by nearly all applicable
records; (ii)~\emph{support}, whether the relation applies to enough records to
be nontrivial---for categorical logical dependencies, each determinant
configuration represented by a rule must occur in sufficiently many rows; and
(iii)~\emph{semantic meaningfulness}, whether the constraint captures a
substantive cross-column relation rather than an accidental regularity. We
retain a candidate only if it satisfies all three criteria. This process yields
42 ground-truth constraints: 13 logical dependencies for \dataset{Anxiety}, 12
equations for \dataset{NBA}, and 17 linear inequalities for \dataset{URL}.
Tables~\ref{tab:anxiety-ground-truth}--\ref{tab:url-ground-truth} list the
complete sets. These annotations remain hidden from all discovery methods and
are used only for evaluation.

\paragraph{\dataset{Anxiety}: logical dependencies.}
We use \dataset{Anxiety} for logical-dependency evaluation because its
low-cardinality categorical view contains semantically ordered demographic,
lifestyle, physiological, and clinical bands. These properties naturally give
rise to conditional admissibility rules---for example, a particular anxiety
band restricting the allowed values of another band---and to mappings between
original and derived categorical attributes.

The \dataset{Anxiety} rules are meaningful because they encode admissible
categories for the subgroup with the highest anxiety score, rather than claiming
that the same associations hold for every participant.

See Table~\ref{tab:anxiety-ground-truth} for the full list of annotated logical dependency ground truths. The first eight high-anxiety rules categoricalize prior expert annotations; the
sleep, stress, and sweating candidates were retained only after independent
semantic review and zero-violation held-out validation; and the final two rules
encode the one-to-many implications among two caffeine consumption-related columns. 

\paragraph{\dataset{NBA}: equational constraints.}
We use \dataset{NBA} for equational-constraint evaluation because its numerical
columns are per-100-possession basketball statistics, many of which are
deterministically derived from shared event counts. Points, makes, attempts,
assisted scoring, and percentages must therefore satisfy exact accounting
identities rather than merely exhibit strong correlations. The annotators
proposed and verified these rules with reference to the NBA Stats glossary for
standard basketball terminology and statistical
definitions.\footnote{\url{https://www.nba.com/stats/help/glossary}}

See Table~\ref{tab:nba-ground-truth} for the full list of annotated equations. 
In the column names, \texttt{FG} denotes field goal; \texttt{2} and \texttt{3}
distinguish two- and three-point shots; \texttt{M} and \texttt{A} denote made
and attempted; \texttt{Pts} denotes points; and \texttt{Pct} denotes a
percentage represented as a fraction on a zero-to-one scale. All counting
statistics are normalized per 100 offensive possessions. Equations 1-3
equations conserve points by decomposing them by shot value and assisted
status; Equations~4--10 reconstruct makes, attempt shares, effective field-goal
percentage, and assisted scoring from their defining numerators and
denominators; and Equations~11--12 reconstruct blocked-attempt counts from their
rates. Changing one participating value while holding the others fixed thus
creates an internally inconsistent player-season record. We require finite
numeric values and test equality with absolute and relative tolerances of
$10^{-9}$, which prevents floating-point representation from turning an
otherwise valid identity into a violation.

\paragraph{\dataset{URL}: linear inequalities.}
We use \dataset{URL} for linear-inequality evaluation because its numerical
features include nested string lengths, delimiter and resource counts, token
order statistics, and complementary ratios. Containment, ordering, and bounded
totals impose natural linear upper and lower bounds on these quantities, making
the dataset especially well suited to semantically verifiable inequalities.

See Table~\ref{tab:url-ground-truth} for the full list of annotated linear inequalities. 
\input{ground_truth_constraint_tables}

In the feature names, \texttt{length\_url} and \texttt{length\_hostname}
denote character lengths, whereas \texttt{length\_words\_raw} counts raw
tokens; \texttt{nb} denotes a count; \texttt{host} and \texttt{path} identify
URL components; \texttt{raw} refers to tokens from the complete URL;
\texttt{shortest}, \texttt{avg}, and \texttt{longest} summarize token lengths;
and \texttt{int} and \texttt{ext} denote internal and external resources.
Hyperlink ratios are fractions on a zero-to-one scale, whereas media ratios are
percentages on a zero-to-100 scale. Feature names follow the source schema;
\texttt{nb\_semicolumn} is its original spelling. A component, token, or count
cannot exceed the URL or hostname that contains it (Constraints~1--8 and 17),
and the minimum, mean, and maximum token lengths must occur in that order
(Constraints~9--14). Internal and external hyperlinks form disjoint fractions
whose sum is at most one, while the corresponding media features sum to at
most 100 (Constraints~15--16). 

\subsection{Detection Evaluation Protocol}

\paragraph{Discovery--audit splits.}
For each dataset, we generate five 70/30 discovery--audit splits using
different random seeds. Only the discovery partition is exposed to the
constraint-discovery method; the audit partition and all ground-truth
annotations remain hidden during discovery. Our agentic
system examines two disjoint samples of 100 rows, perform three discovery runs
and verifies them against the complete discovery partition,
and may revise failed proposals for up to three rounds using verifier-provided
counterexamples. Their naive baselines receive the same total sample budget of
200 rows in a single prompt and have no access
to verification tools or iterative revision. 

\paragraph{Contrastive detection tasks.}
Every example begins with an audit row that satisfies all ground-truth
constraints and pairs it with a counterpart obtained by changing exactly one
cell. For \dataset{URL}, we move a participating value across the target half-space
boundary; each of the 17 inequalities receives 1,000 matched pairs. For
\dataset{NBA},
we perturb a target-specific numerical column using noise scaled by the
discovery-column standard deviation; each of the 12 equations also receives
1,000 pairs. We retain a mutation only when it violates the target constraint
while satisfying every other ground-truth constraint. The \dataset{URL} and
\dataset{NBA} tasks
therefore contain 34,000 and 24,000 shuffled rows per split, respectively.

For \dataset{Anxiety}, task allocation is proportional to each rule's full-data
applicability support while preserving a budget of 1,950 matched pairs per
split. Each of the 11 highest-anxiety implications applies to 651 of 11,000
source rows and receives 51 pairs. The occupation-to-group mapping applies to
all 11,000 rows and receives 861 pairs, while the occupation-group-to-caffeine
mapping applies to 6,751 rows and receives 528 pairs. A mutation changes only
the dependent column to a disallowed category and is retained only when the
target mapping fails and all other mappings still pass. Every
constraint-specific task remains balanced between valid and invalid rows;
support weighting captures how often a rule applies, not the natural
prevalence of violations. Across the five splits, the three benchmarks contain
85,000, 60,000, and 9,750 matched pairs, respectively. Audit rows may be reused
across constraint-specific tasks, so these totals count evaluation examples
rather than unique source records.

\paragraph{Prediction and metric aggregation.}
We execute the constraints discovered by each method directly as an ensemble:
a row is classified as invalid if any discovered constraint rejects it. We do
not manually match discovered constraints to ground-truth formulas or revise
them during evaluation. Malformed or non-executable constraints produce no
violation predictions. We compute accuracy, precision, and recall for each
detection task and take a support-weighted macro-average across the constraints
of a given dataset. For metric $m$ in a split, we report
$\sum_c n_c m_c / \sum_c n_c$, where $m_c$ is the metric for constraint $c$
and $n_c$ is its allocated number of matched audit pairs. This allocation is
proportional to full-data applicability support for \dataset{Anxiety}; the
weights are equal for \dataset{URL} and \dataset{NBA} because every constraint
receives 1,000 pairs. We report the mean and standard deviation of these
weighted scores across the five discovery--audit splits. This protocol evaluates behavioral
detection rather than exact formula recovery: a discovered constraint receives
credit when it detects held-out violations even if its syntax differs from the
corresponding curated rule.

%% file: ground_truth_constraint_tables.tex
\begin{table*}[p]
  \centering
  \footnotesize
  \setlength{\tabcolsep}{4pt}
  \renewcommand{\arraystretch}{1.08}
  \begin{tabular}{@{}r>{\raggedright\arraybackslash}p{0.92\textwidth}@{}}
    \toprule
    \# & Ground-truth logical dependency \\
    \midrule
    1 & \texttt{Anxiety Band} = \texttt{Very high (9--10)} $\rightarrow$
        \texttt{Caffeine Band} = \texttt{294 mg or more} \\
    2 & \texttt{Anxiety Band} = \texttt{Very high (9--10)} $\rightarrow$
        \texttt{Breathing Rate Band} = \texttt{20 breaths/min or higher} \\
    3 & \texttt{Anxiety Band} = \texttt{Very high (9--10)} $\rightarrow$
        \texttt{Heart Rate Band} = \texttt{85 bpm or higher} \\
    4 & \texttt{Anxiety Band} = \texttt{Very high (9--10)} $\rightarrow$
        \texttt{Alcohol Band} = \texttt{5 or more drinks} \\
    5 & \texttt{Anxiety Band} = \texttt{Very high (9--10)} $\rightarrow$
        \texttt{Diet Quality Band} = \texttt{Low (1--4)} \\
    6 & \texttt{Anxiety Band} = \texttt{Very high (9--10)} $\rightarrow$
        \texttt{Therapy Band} = \texttt{3 or more sessions} \\
    7 & \texttt{Anxiety Band} = \texttt{Very high (9--10)} $\rightarrow$
        \texttt{Age Band} = \texttt{20--49} \\
    8 & \texttt{Anxiety Band} = \texttt{Very high (9--10)} $\rightarrow$
        \texttt{Physical Activity Band} = \texttt{0--4 hours} \\
    9 & \texttt{Anxiety Band} = \texttt{Very high (9--10)} $\rightarrow$
        \texttt{Sleep Band} = \{\texttt{Under 6 hours},
        \texttt{6--9 hours}\} \\
    10 & \texttt{Anxiety Band} = \texttt{Very high (9--10)} $\rightarrow$
         \texttt{Stress Band} = \{\texttt{High (7--8)},
         \texttt{Very high (9--10)}\} \\
    11 & \texttt{Anxiety Band} = \texttt{Very high (9--10)} $\rightarrow$
         \texttt{Sweating Band} = \{\texttt{Moderate (3)},
         \texttt{High (4--5)}\} \\
    12 & \texttt{Occupation} = \{\texttt{Scientist}, \texttt{Doctor},
         \texttt{Engineer}, \texttt{Lawyer}\} $\rightarrow$
         \texttt{Occupation Group} = \texttt{Higher caffeine floor};
         \texttt{Occupation} = \{\texttt{Student}, \texttt{Nurse},
         \texttt{Freelancer}, \texttt{Chef}\} $\rightarrow$
         \texttt{Occupation Group} = \texttt{Moderate caffeine floor};
         \texttt{Occupation} = \{\texttt{Artist}, \texttt{Athlete},
         \texttt{Musician}, \texttt{Other}, \texttt{Teacher}\} $\rightarrow$
         \texttt{Occupation Group} = \texttt{No annotated caffeine floor} \\
    13 & \texttt{Occupation Group} = \texttt{Higher caffeine floor}
         $\rightarrow$ \texttt{Caffeine Band} = \{\texttt{250--293 mg},
         \texttt{294 mg or more}\}; \texttt{Occupation Group} =
         \texttt{Moderate caffeine floor} $\rightarrow$
         \texttt{Caffeine Band} = \{\texttt{100--249 mg},
         \texttt{250--293 mg}, \texttt{294 mg or more}\} \\
    \bottomrule
  \end{tabular}
  \caption{The 13 annotator-verified logical dependencies used as ground truth for
  the \dataset{Anxiety} detection task.}
  \label{tab:anxiety-ground-truth}
\vspace{1em}

  \centering
  \footnotesize
  \setlength{\tabcolsep}{4pt}
  \renewcommand{\arraystretch}{1.08}
  \begin{tabular}{@{}r>{\raggedright\arraybackslash}p{0.92\textwidth}@{}}
    \toprule
    \# & Ground-truth equation \\
    \midrule
    1 & $\colname{Points}=2\times\colname{FG2M}+3\times\colname{FG3M}+\colname{FtPoints}$ \\
    2 & $2\times\colname{FG2M}=\colname{PtsAssisted2s}+\colname{PtsUnassisted2s}$ \\
    3 & $3\times\colname{FG3M}=\colname{PtsAssisted3s}+\colname{PtsUnassisted3s}$ \\
    4 & $\colname{Fg2Pct}\times\colname{FG2A}=\colname{FG2M}$ \\
    5 & $\colname{Fg3Pct}\times\colname{FG3A}=\colname{FG3M}$ \\
    6 & $\colname{FG3APct}\times(\colname{FG2A}+\colname{FG3A})=\colname{FG3A}$ \\
    7 & $\colname{EfgPct}\times(\colname{FG2A}+\colname{FG3A})
         =\colname{FG2M}+1.5\times\colname{FG3M}$ \\
    8 & $\colname{Assisted2sPct}\times(2\times\colname{FG2M})=\colname{PtsAssisted2s}$ \\
    9 & $\colname{Assisted3sPct}\times(3\times\colname{FG3M})=\colname{PtsAssisted3s}$ \\
    10 & $\colname{NonPutbacksAssisted2sPct}
          \times(2\times\colname{FG2M}-\colname{PtsPutbacks})=\colname{PtsAssisted2s}$ \\
    11 & $\colname{FG2APctBlocked}\times\colname{FG2A}=\colname{Fg2aBlocked}$ \\
    12 & $\colname{FG3APctBlocked}\times\colname{FG3A}=\colname{Fg3aBlocked}$ \\
    \bottomrule
  \end{tabular}
  \caption{The 12 annotator-verified equations used as ground truth for the
  \dataset{NBA} detection task.}
  \label{tab:nba-ground-truth}
\vspace{1em}

  \centering
  \footnotesize
  \setlength{\tabcolsep}{4pt}
  \renewcommand{\arraystretch}{1.08}
  \begin{tabular}{@{}r>{\raggedright\arraybackslash}p{0.92\textwidth}@{}}
    \toprule
    \# & Ground-truth linear inequality \\
    \midrule
    1 & $\colname{length_url}-\colname{length_hostname}
         -\colname{longest_word_path}-\colname{nb_slash}\geq 0$ \\
    2 & $\colname{length_url}-\colname{nb_dots}-\colname{nb_hyphens}
         -\colname{nb_and}-\colname{nb_eq}-\colname{nb_underscore}
         -\colname{nb_percent}-\colname{nb_slash}-\colname{nb_semicolumn}\geq 0$ \\
    3 & $\colname{length_url}-\colname{length_words_raw}\geq 0$ \\
    4 & $\colname{length_url}-\colname{longest_words_raw}\geq 0$ \\
    5 & $\colname{length_hostname}-\colname{longest_word_host}\geq 0$ \\
    6 & $\colname{longest_words_raw}-\colname{longest_word_host}\geq 0$ \\
    7 & $\colname{longest_words_raw}-\colname{longest_word_path}\geq 0$ \\
    8 & $\colname{shortest_word_host}-\colname{shortest_words_raw}\geq 0$ \\
    9 & $\colname{longest_words_raw}-\colname{avg_words_raw}\geq 0$ \\
    10 & $\colname{avg_words_raw}-\colname{shortest_words_raw}\geq 0$ \\
    11 & $\colname{longest_word_host}-\colname{avg_word_host}\geq 0$ \\
    12 & $\colname{avg_word_host}-\colname{shortest_word_host}\geq 0$ \\
    13 & $\colname{longest_word_path}-\colname{avg_word_path}\geq 0$ \\
    14 & $\colname{avg_word_path}-\colname{shortest_word_path}\geq 0$ \\
    15 & $-\colname{ratio_intHyperlinks}-\colname{ratio_extHyperlinks}\geq -1$ \\
    16 & $-\colname{ratio_intMedia}-\colname{ratio_extMedia}\geq -100$ \\
    17 & $\colname{nb_hyperlinks}-\colname{nb_extCSS}\geq 0$ \\
    \bottomrule
  \end{tabular}
  \caption{The 17 annotator-verified linear inequalities used as ground truth for
  the \dataset{URL} detection task.}
  \label{tab:url-ground-truth}
\end{table*}

%% file: end_to_end_constraint_aware_generation.tex
\section{End-to-End Constraint-Aware Generation Details}
\label{app:end_to_end_details}

\subsection{Generator Training Details}

Tables~\ref{tab:ctgan-configuration}--\ref{tab:tabddpm-configuration} report
the resolved hyperparameter settings used for the four generators. Our CTGAN
and TVAE implementations are
adapted from the implementations released in the official code repository of
\citet{stoian2024how}.\footnote{\url{https://github.com/mihaela-stoian/ConstrainedDGM}}
Our TabDDPM implementation is adapted from the official implementation of
\citet{kotelnikov2023tabddpm}.\footnote{\url{https://github.com/yandex-research/tab-ddpm}}
For Gaussian Copula, we use SDV~1.32.1's
\texttt{GaussianCopulaSynthesizer}.\footnote{\url{https://docs.sdv.dev/sdv/modeling/single-table-synthesizers/gaussiancopulasynthesizer}}

\begin{table}[H]
  \centering
  \footnotesize
  \setlength{\tabcolsep}{4pt}
  \renewcommand{\arraystretch}{1.06}
  \begin{tabular}{@{}>{\raggedright\arraybackslash}p{0.56\columnwidth}
                      >{\raggedright\arraybackslash}p{0.34\columnwidth}@{}}
    \toprule
    \textbf{CTGAN hyperparameter} & \textbf{Setting} \\
    Training epochs & 300 \\
    Batch size & 500 \\
    Noise/embedding dimension & 128 \\
    Generator hidden layers & $[256,256]$ \\
    Discriminator hidden layers & $[256,256]$ \\
    Generator learning rate & $2\times10^{-4}$ \\
    Discriminator learning rate & $2\times10^{-4}$ \\
    Generator weight decay & $1\times10^{-6}$ \\
    Discriminator weight decay & $1\times10^{-6}$ \\
    PacGAN packing size (\texttt{pac}) & 10 \\
    Optimizer & Adam \\
    Gumbel--Softmax temperature & 0.2 \\
    \bottomrule
  \end{tabular}
  \caption{CTGAN training configuration.}
  \label{tab:ctgan-configuration}
\end{table}

\begin{table}[H]
  \centering
  \footnotesize
  \setlength{\tabcolsep}{4pt}
  \renewcommand{\arraystretch}{1.06}
  \begin{tabular}{@{}>{\raggedright\arraybackslash}p{0.56\columnwidth}
                      >{\raggedright\arraybackslash}p{0.34\columnwidth}@{}}
    \toprule
    \textbf{TVAE hyperparameter} & \textbf{Setting} \\
    Training epochs & 300 \\
    Batch size & 500 \\
    Latent/embedding dimension & 128 \\
    Encoder/compression layers & $[128,128]$ \\
    Decoder/decompression layers & $[128,128]$ \\
    Optimizer & Adam \\
    Learning rate & $1\times10^{-3}$ \\
    \bottomrule
  \end{tabular}
  \caption{TVAE training configuration.}
  \label{tab:tvae-configuration}
\end{table}

\begin{table}[H]
  \centering
  \footnotesize
  \setlength{\tabcolsep}{4pt}
  \renewcommand{\arraystretch}{1.06}
  \begin{tabular}{@{}>{\raggedright\arraybackslash}p{0.56\columnwidth}
                      >{\raggedright\arraybackslash}p{0.34\columnwidth}@{}}
    \toprule
    \textbf{Gaussian Copula hyperparameter} & \textbf{Setting} \\
    Default numerical marginal distribution & Beta \\
    \bottomrule
  \end{tabular}
  \caption{Gaussian Copula configuration.}
  \label{tab:gaussian-copula-configuration}
\end{table}

\begin{table}[H]
  \centering
  \footnotesize
  \setlength{\tabcolsep}{4pt}
  \renewcommand{\arraystretch}{1.06}
  \begin{tabular}{@{}>{\raggedright\arraybackslash}p{0.56\columnwidth}
                      >{\raggedright\arraybackslash}p{0.34\columnwidth}@{}}
    \toprule
    \textbf{TabDDPM hyperparameter} & \textbf{Setting} \\
    Optimizer updates & 30,000 \\
    Maximum training batch size & 4,096 \\
    Optimizer & AdamW \\
    Initial learning rate & $1\times10^{-3}$ \\
    Learning-rate schedule & Linear decay toward zero \\
    Denoising MLP hidden layers & $[256,256]$ \\
    Hidden-layer dropout & 0 \\
    Time/label embedding dimension & 128 \\
    Diffusion timesteps & 1,000 \\
    Diffusion $\beta$ schedule & Cosine \\
    Numerical loss & MSE \\
    EMA decay & 0.999 \\
    \multicolumn{2}{@{}l@{}}{\textbf{\dataset{Flights}-only exception}} \\
    Hidden layers & $[256,512,512,256]$ \\
    Diffusion timesteps & 100 \\
    \bottomrule
  \end{tabular}
  \caption{TabDDPM training configuration, including the
  \dataset{Flights}-specific override.}
  \label{tab:tabddpm-configuration}
\end{table}

\subsection{Definition of Metrics}
\paragraph{Constraint metrics.}
Constraint violation rate (CVR) and sample-wise constraint violation coverage
(sCVC) are common constraint metrics \citep{stoian2025survey}.
Let $n$ be the number of synthetic rows, $\mathcal{C}_f$ the constraints in
family $f$, and $v_{ic}\in\{0,1\}$ indicate whether row $i$ violates
constraint $c$. We define
\begin{equation*}
  \operatorname{CVR}_f
  =\frac{1}{n}\sum_{i=1}^{n}
    \mathbb{I}\!\left[\sum_{c\in\mathcal{C}_f}v_{ic}>0\right].
\end{equation*}
\begin{equation*}
  \operatorname{sCVC}_f
  =\frac{1}{n|\mathcal{C}_f|}\sum_{i=1}^{n}
    \sum_{c\in\mathcal{C}_f}v_{ic}.
\end{equation*}
CVR is the fraction of rows that violate at least one constraint, while sCVC is
the average fraction of constraints violated per row. Both metrics lie in
$[0,1]$, zero indicating no measured violations under the evaluated constraints. As the number
of constraints grows, CVR can easily approach one because any single violation
marks the entire row as violating; sCVC remains more informative in this regime
because it measures violation density over all row--constraint checks.

For each evaluated equational constraint $c$ over columns $S_c$, let the
nonempty set $\mathcal{T}_c\subseteq S_c$ contain exactly the verified repair
targets. A repair derives a target column $j\in\mathcal{T}_c$ from the remaining
involved columns $S_c\setminus\{j\}$. We use the validated reconstruction
function
$g_{c\rightarrow j}$ implemented by the generated \texttt{fix(df)} code in the
equational-constraint enforcement procedure described in
Section~\ref{sec:constraint_enforcement}. The evaluator considers every
verified target direction and retains the best coefficient of determination.
Equational consistency is
\begin{equation*}
  R^2_{\mathrm{eq}}
  =\frac{1}{|\mathcal{C}_{\mathrm{eq}}|}
   \sum_{c\in\mathcal{C}_{\mathrm{eq}}}
   \max_{j\in\mathcal{T}_c}R^2\!\left(x_j,
   g_{c\rightarrow j}(x_{S_c\setminus\{j\}})\right).
\end{equation*}
Higher values are better, and one indicates exact consistency with every
evaluated equation.

Let $\mathcal{F}$ be the joint feasible region of the discovered linear
constraints, and let $s_j$ be the population standard deviation of numerical
column $j$ in the real training data. To keep the normalization defined for
constant columns, set $\widetilde{s}_j=s_j$ when $s_j>0$ and
$\widetilde{s}_j=1$ when $s_j=0$, and let
$\widetilde{D}=\operatorname{diag}((\widetilde{s}_j)_{j\in S_{\mathrm{num}}})$,
where $S_{\mathrm{num}}$ indexes the numerical columns.
Linear feasibility distance is
\begin{equation*}
  \operatorname{LFD}
  =\frac{1}{n}\sum_{i=1}^{n}
    \min_{z\in\mathcal{F}}\left\|D^{-1}(x_i-z)\right\|_2.
\end{equation*}
Lower values are better, and zero indicates that every row lies in the joint
feasible region.

\paragraph{Utility and univariate marginal fidelity.}
Under Train on Synthetic, Test on Real (TSTR) \citep{stoian2025survey}, a
predictor $h_{\mathrm{syn}}$ is fitted to a synthetic table and evaluated on
the corresponding real test partition:
\begin{equation*}
  \mathrm{TSTR}
  =\operatorname{Score}\!\left(
    h_{\mathrm{syn}}=\operatorname{Train}(\mathcal{D}_{\mathrm{syn}}),
    \mathcal{D}_{\mathrm{test}}^{\mathrm{real}}\right).
\end{equation*}
We report ROC-AUC for classification and $R^2$ for regression. Candidate
classification models are logistic regression, decision tree, and XGBoost;
candidate regression models are linear regression and XGBoost. For each
dataset, we select the model with the highest mean Train on Real, Test on Real (TRTR)
score. This selects logistic regression for
\dataset{HELOC}; linear regression for \dataset{Flights} and \dataset{NBA}; and
XGBoost for \dataset{News}, \dataset{Steel}, \dataset{Taxi}, and \dataset{URL}.

Column Shapes is the mean univariate similarity over the $d$ columns. For a
numerical column $j$, SDMetrics uses the KS complement; for a categorical
column, it uses the total-variation complement:
\begin{equation*}
  q_j^{\mathrm{num}}
    =1-\sup_z\left|\widehat F_{\mathrm{real}}^j(z)
                       -\widehat F_{\mathrm{syn}}^j(z)\right|.
\end{equation*}
\begin{equation*}
  q_j^{\mathrm{cat}}
    =1-\frac{1}{2}\sum_a
      \left|p_{\mathrm{real}}^j(a)-p_{\mathrm{syn}}^j(a)\right|.
\end{equation*}
\begin{equation*}
  \operatorname{Column \ Shapes}
    =\frac{1}{d}\sum_{j=1}^{d}q_j.
\end{equation*}
All three scores lie between zero and one, with higher values indicating more
similar marginal distributions.

\subsection{Computational resources}
Our experiments were conducted on a  system equipped with two
64-core AMD EPYC 7662 processors (128 CPU cores in total) and four NVIDIA RTX
6000 Ada Generation GPUs, each with 48~GB of memory. The full design comprised
84 dataset--split--synthesizer settings ($7$ datasets $\times$ $3$ splits
$\times$ $4$ synthesizers) and 252 synthetic-sample runs (three per setting).
The recorded training and generation stages required approximately 18 hours of cumulative computation. If distributed well across all four GPUs, the workloads
would require approximately 5 hours.

\subsection{Full End-to-End Results}
\label{app:end_to_end_full_results}

Table~\ref{tab:full-statistical-fidelity-results} reports Column Shapes for
every dataset, generator, and output variant; higher values indicate greater
univariate marginal fidelity to the real data.

\input{statistic_results.tex}

Table~\ref{tab:full-utility-results} reports the complete utility results,
including the selected dataset-level TRTR result and the raw and
constraint-postprocessed TSTR score for every dataset--generator pair.

\input{utility_results}

Table~\ref{tab:full-raw-constraint-results} reports every raw
dataset--generator constraint result. We report only raw scores because
postprocessing addresses the applicable constraints effectively and attains
perfect constraint scores throughout: CVR, sCVC, and LFD are zero, while
equational $R^2$ is one.

\input{raw_constraint_results}

\FloatBarrier

%% file: statistic_results.tex
\begin{table}[H]
    \centering
    \scriptsize
    \setlength{\tabcolsep}{1.5pt}
    \renewcommand{\arraystretch}{0.94}
    \begin{tabular}{@{}llccc@{}}
      \toprule
      & & \multicolumn{3}{c}{\textbf{Column Shapes $\uparrow$}} \\
      \cmidrule(l){3-5}
      \textbf{Dataset} & \textbf{Generator} & \textbf{Raw} &
      \textbf{Postprocessed} & \textbf{$\Delta$} \\
      \midrule
      \multirow{4}{*}{\dataset{Flights}}
        & CTGAN & $0.906\;(0.008)$ & $0.912\;(0.010)$ & \deltapos{+0.006} \\
        & Gaussian Copula & $0.931\;(0.001)$ & $0.931\;(0.001)$ & \deltazero \\
        & TabDDPM & $0.968\;(0.004)$ & $0.969\;(0.003)$ & \deltapos{+0.001} \\
        & TVAE & $0.932\;(0.002)$ & $0.935\;(0.004)$ & \deltapos{+0.003} \\
      \midrule
      \multirow{4}{*}{\dataset{HELOC}}
        & CTGAN & $0.906\;(0.014)$ & $0.900\;(0.016)$ & \deltaneg{-0.006} \\
        & Gaussian Copula & $0.836\;(0.021)$ & $0.811\;(0.033)$ & \deltaneg{-0.025} \\
        & TabDDPM & $0.850\;(0.038)$ & $0.850\;(0.038)$ & \deltazero \\
        & TVAE & $0.930\;(0.006)$ & $0.927\;(0.005)$ & \deltaneg{-0.003} \\
      \midrule
      \multirow{4}{*}{\dataset{NBA}}
        & CTGAN & $0.889\;(0.019)$ & $0.893\;(0.012)$ & \deltapos{+0.004} \\
        & Gaussian Copula & $0.824\;(0.016)$ & $0.846\;(0.014)$ & \deltapos{+0.022} \\
        & TabDDPM & $0.971\;(0.009)$ & $0.982\;(0.001)$ & \deltapos{+0.011} \\
        & TVAE & $0.896\;(0.004)$ & $0.897\;(0.007)$ & \deltapos{+0.001} \\
      \midrule
      \multirow{4}{*}{\dataset{News}}
        & CTGAN & $0.862\;(0.006)$ & $0.854\;(0.008)$ & \deltaneg{-0.008} \\
        & Gaussian Copula & $0.832\;(0.005)$ & $0.829\;(0.003)$ & \deltaneg{-0.003} \\
        & TabDDPM & $0.544\;(0.041)$ & $0.541\;(0.029)$ & \deltaneg{-0.003} \\
        & TVAE & $0.871\;(0.009)$ & $0.862\;(0.010)$ & \deltaneg{-0.009} \\
      \midrule
      \multirow{4}{*}{\dataset{Steel}}
        & CTGAN & $0.890\;(0.015)$ & $0.900\;(0.012)$ & \deltapos{+0.010} \\
        & Gaussian Copula & $0.816\;(0.010)$ & $0.861\;(0.010)$ & \deltapos{+0.045} \\
        & TabDDPM & $0.985\;(0.001)$ & $0.985\;(0.001)$ & \deltazero \\
        & TVAE & $0.916\;(0.032)$ & $0.937\;(0.016)$ & \deltapos{+0.021} \\
      \midrule
      \multirow{4}{*}{\dataset{Taxi}}
        & CTGAN & $0.954\;(0.004)$ & $0.956\;(0.005)$ & \deltapos{+0.002} \\
        & Gaussian Copula & $0.856\;(0.032)$ & $0.863\;(0.027)$ & \deltapos{+0.007} \\
        & TabDDPM & $0.982\;(0.012)$ & $0.982\;(0.012)$ & \deltazero \\
        & TVAE & $0.964\;(0.002)$ & $0.965\;(0.002)$ & \deltapos{+0.001} \\
      \midrule
      \multirow{4}{*}{\dataset{URL}}
        & CTGAN & $0.895\;(0.019)$ & $0.898\;(0.018)$ & \deltapos{+0.003} \\
        & Gaussian Copula & $0.862\;(0.017)$ & $0.861\;(0.016)$ & \deltaneg{-0.001} \\
        & TabDDPM & $0.584\;(0.011)$ & $0.584\;(0.013)$ & \deltazero \\
        & TVAE & $0.937\;(0.001)$ & $0.934\;(0.004)$ & \deltaneg{-0.003} \\
      \bottomrule
    \end{tabular}
    \caption{Column Shapes for all dataset--generator combinations before and
    after constraint postprocessing. Entries are means, with standard
    deviations in parentheses; each entry aggregates nine runs across three
    independent data splits. The final column gives
    $\Delta=\text{Postprocessed}-\text{Raw}$ for the displayed means; green and
    orange denote gains and losses, respectively. Higher is better.}
    \label{tab:full-statistical-fidelity-results}
\end{table}

%% file: utility_results.tex
\begin{table*}[t]
  \centering
  \scriptsize
  \setlength{\tabcolsep}{2pt}
  \renewcommand{\arraystretch}{0.98}
  \begin{tabular}{@{}llllcccc@{}}
    \toprule
    \textbf{Dataset} & \textbf{Task / metric} &
    \textbf{Selected evaluation model} & \textbf{Generator} &
    \textbf{TRTR} $\uparrow$ & \textbf{Raw TSTR} $\uparrow$ &
    \textbf{Postprocessed TSTR} $\uparrow$ & \textbf{$\Delta$} \\
    \midrule
    \multirow{4}{*}{\dataset{Flights}} & \multirow{4}{*}{Regression ($R^2$)} & \multirow{4}{*}{Linear regression} & CTGAN & \multirow{4}{*}{$0.986\;(0.000)$} & $0.645\;(0.037)$ & $0.661\;(0.040)$ & \deltapos{+0.016} \\
     & & & Gaussian Copula & & $0.963\;(0.008)$ & $0.964\;(0.002)$ & \deltapos{+0.001} \\
     & & & TabDDPM & & $0.839\;(0.052)$ & $0.745\;(0.111)$ & \deltaneg{-0.094} \\
     & & & TVAE & & $0.875\;(0.078)$ & $0.874\;(0.077)$ & \deltaneg{-0.001} \\
    \midrule
    \multirow{4}{*}{\dataset{HELOC}} & \multirow{4}{*}{Classification (ROC-AUC)} & \multirow{4}{*}{Logistic regression} & CTGAN & \multirow{4}{*}{$0.802\;(0.003)$} & $0.720\;(0.028)$ & $0.720\;(0.029)$ & \deltazero \\
     & & & Gaussian Copula & & $0.775\;(0.020)$ & $0.784\;(0.004)$ & \deltapos{+0.009} \\
     & & & TabDDPM & & $0.775\;(0.009)$ & $0.777\;(0.008)$ & \deltapos{+0.002} \\
     & & & TVAE & & $0.790\;(0.002)$ & $0.790\;(0.002)$ & \deltazero \\
    \midrule
    \multirow{4}{*}{\dataset{NBA}} & \multirow{4}{*}{Regression ($R^2$)} & \multirow{4}{*}{Linear regression} & CTGAN & \multirow{4}{*}{$0.938\;(0.014)$} & $0.712\;(0.043)$ & $0.614\;(0.092)$ & \deltaneg{-0.098} \\
     & & & Gaussian Copula & & $0.923\;(0.014)$ & $0.892\;(0.044)$ & \deltaneg{-0.031} \\
     & & & TabDDPM & & $0.885\;(0.011)$ & $0.900\;(0.008)$ & \deltapos{+0.015} \\
     & & & TVAE & & $0.902\;(0.021)$ & $0.845\;(0.038)$ & \deltaneg{-0.057} \\
    \midrule
    \multirow{4}{*}{\dataset{News}} & \multirow{4}{*}{Classification (ROC-AUC)} & \multirow{4}{*}{XGBoost} & CTGAN & \multirow{4}{*}{$0.726\;(0.002)$} & $0.634\;(0.021)$ & $0.628\;(0.021)$ & \deltaneg{-0.006} \\
     & & & Gaussian Copula & & $0.589\;(0.008)$ & $0.588\;(0.010)$ & \deltaneg{-0.001} \\
     & & & TabDDPM & & $0.668\;(0.003)$ & $0.668\;(0.006)$ & \deltazero \\
     & & & TVAE & & $0.628\;(0.014)$ & $0.620\;(0.014)$ & \deltaneg{-0.008} \\
    \midrule
    \multirow{4}{*}{\dataset{Steel}} & \multirow{4}{*}{Regression ($R^2$)} & \multirow{4}{*}{XGBoost} & CTGAN & \multirow{4}{*}{$0.999\;(0.000)$} & $0.880\;(0.010)$ & $0.996\;(0.001)$ & \deltapos{+0.116} \\
     & & & Gaussian Copula & & $-0.850\;(1.011)$ & $0.914\;(0.113)$ & \deltapos{+1.764} \\
     & & & TabDDPM & & $0.986\;(0.001)$ & $0.998\;(0.000)$ & \deltapos{+0.012} \\
     & & & TVAE & & $0.930\;(0.016)$ & $0.997\;(0.000)$ & \deltapos{+0.067} \\
    \midrule
    \multirow{4}{*}{\dataset{Taxi}} & \multirow{4}{*}{Regression ($R^2$)} & \multirow{4}{*}{XGBoost} & CTGAN & \multirow{4}{*}{$0.801\;(0.004)$} & $0.625\;(0.040)$ & $0.647\;(0.033)$ & \deltapos{+0.022} \\
     & & & Gaussian Copula & & $0.633\;(0.042)$ & $0.629\;(0.046)$ & \deltaneg{-0.004} \\
     & & & TabDDPM & & $0.447\;(0.184)$ & $0.491\;(0.050)$ & \deltapos{+0.044} \\
     & & & TVAE & & $0.729\;(0.044)$ & $0.738\;(0.042)$ & \deltapos{+0.009} \\
    \midrule
    \multirow{4}{*}{\dataset{URL}} & \multirow{4}{*}{Classification (ROC-AUC)} & \multirow{4}{*}{XGBoost} & CTGAN & \multirow{4}{*}{$0.994\;(0.000)$} & $0.840\;(0.034)$ & $0.839\;(0.019)$ & \deltaneg{-0.001} \\
     & & & Gaussian Copula & & $0.903\;(0.005)$ & $0.907\;(0.011)$ & \deltapos{+0.004} \\
     & & & TabDDPM & & $0.487\;(0.095)$ & $0.541\;(0.057)$ & \deltapos{+0.054} \\
     & & & TVAE & & $0.966\;(0.006)$ & $0.965\;(0.007)$ & \deltaneg{-0.001} \\
    \bottomrule
  \end{tabular}
  \caption{Complete downstream utility results (mean; standard deviation in
  parentheses). TRTR aggregates three splits for the selected dataset-level
  model; TSTR aggregates nine runs across three splits for each
  dataset--generator pair. The final column reports
  $\Delta=\text{Postprocessed}-\text{Raw}$ for displayed TSTR means (green:
  gain; orange: loss). Regression uses $R^2$ and classification uses ROC-AUC;
  higher is better.}
  \label{tab:full-utility-results}
\end{table*}

%% file: raw_constraint_results.tex
\begin{table*}[t]
  \centering
  \scriptsize
  \setlength{\tabcolsep}{1.7pt}
  \renewcommand{\arraystretch}{1.03}
  \resizebox{\textwidth}{!}{%
  \begin{tabular}{@{}llcccccccc@{}}
    \toprule
    & & \multicolumn{2}{c}{\textbf{Logical dependency}} &
    \multicolumn{3}{c}{\textbf{Equational}} &
    \multicolumn{3}{c}{\textbf{Linear}} \\
    \cmidrule(lr){3-4}\cmidrule(lr){5-7}\cmidrule(l){8-10}
    \textbf{Dataset} & \textbf{Generator} &
    \textbf{CVR} $\downarrow$ & \textbf{sCVC} $\downarrow$ &
    \textbf{CVR} $\downarrow$ & \textbf{sCVC} $\downarrow$ &
    $\boldsymbol{R^2}$ $\uparrow$ & \textbf{CVR} $\downarrow$ &
    \textbf{sCVC} $\downarrow$ & \textbf{LFD} $\downarrow$ \\
    \midrule
    \multirow{4}{*}{\dataset{Flights}} & CTGAN & $0.826\;(0.021)$ & $0.455\;(0.013)$ & $1.000\;(0.000)$ & $0.996\;(0.000)$ & $0.301\;(0.031)$ & $0.000\;(0.000)$ & $0.000\;(0.000)$ & $0.000\;(0.000)$ \\
     & Gaussian Copula & $0.994\;(0.002)$ & $0.757\;(0.029)$ & $1.000\;(0.000)$ & $0.998\;(0.000)$ & $-0.214\;(0.005)$ & $0.000\;(0.000)$ & $0.000\;(0.000)$ & $0.000\;(0.000)$ \\
     & TabDDPM & $0.116\;(0.004)$ & $0.057\;(0.005)$ & $1.000\;(0.000)$ & $0.982\;(0.001)$ & $0.691\;(0.033)$ & $0.000\;(0.000)$ & $0.000\;(0.000)$ & $0.000\;(0.000)$ \\
     & TVAE & $0.702\;(0.004)$ & $0.332\;(0.002)$ & $1.000\;(0.000)$ & $0.994\;(0.001)$ & $0.474\;(0.010)$ & $0.000\;(0.000)$ & $0.000\;(0.000)$ & $0.000\;(0.000)$ \\
    \midrule
    \multirow{4}{*}{\dataset{HELOC}} & CTGAN & $0.013\;(0.007)$ & $0.013\;(0.007)$ & -- & -- & -- & $0.715\;(0.051)$ & $0.095\;(0.017)$ & $0.698\;(0.102)$ \\
     & Gaussian Copula & $0.125\;(0.056)$ & $0.125\;(0.056)$ & -- & -- & -- & $0.804\;(0.301)$ & $0.114\;(0.063)$ & $2.254\;(3.121)$ \\
     & TabDDPM & $0.003\;(0.002)$ & $0.003\;(0.002)$ & -- & -- & -- & $0.310\;(0.017)$ & $0.050\;(0.003)$ & $1.462\;(0.645)$ \\
     & TVAE & $0.020\;(0.012)$ & $0.020\;(0.012)$ & -- & -- & -- & $0.519\;(0.021)$ & $0.055\;(0.007)$ & $0.261\;(0.024)$ \\
    \midrule
    \multirow{4}{*}{\dataset{NBA}} & CTGAN & -- & -- & $1.000\;(0.000)$ & $0.955\;(0.029)$ & $-6.140\;(10.488)$ & $0.127\;(0.220)$ & $0.012\;(0.020)$ & $0.076\;(0.132)$ \\
     & Gaussian Copula & -- & -- & $1.000\;(0.000)$ & $1.000\;(0.000)$ & $0.162\;(0.438)$ & $0.082\;(0.141)$ & $0.007\;(0.013)$ & $0.019\;(0.033)$ \\
     & TabDDPM & -- & -- & $0.997\;(0.001)$ & $0.851\;(0.005)$ & $0.302\;(0.310)$ & $0.033\;(0.057)$ & $0.003\;(0.005)$ & $0.001\;(0.002)$ \\
     & TVAE & -- & -- & $1.000\;(0.000)$ & $0.970\;(0.007)$ & $0.475\;(0.139)$ & $0.080\;(0.138)$ & $0.007\;(0.013)$ & $0.019\;(0.033)$ \\
    \midrule
    \multirow{4}{*}{\dataset{News}} & CTGAN & $0.186\;(0.056)$ & $0.019\;(0.003)$ & $1.000\;(0.000)$ & $0.950\;(0.021)$ & $-0.229\;(0.052)$ & $0.413\;(0.578)$ & $0.044\;(0.062)$ & $0.151\;(0.210)$ \\
     & Gaussian Copula & $0.284\;(0.057)$ & $0.032\;(0.004)$ & $1.000\;(0.000)$ & $1.000\;(0.000)$ & $-0.222\;(0.017)$ & $0.455\;(0.634)$ & $0.047\;(0.065)$ & $0.088\;(0.124)$ \\
     & TabDDPM & $0.040\;(0.027)$ & $0.004\;(0.003)$ & $1.000\;(0.000)$ & $0.529\;(0.143)$ & $-0.450\;(0.398)$ & $0.505\;(0.451)$ & $0.106\;(0.105)$ & $5.580\;(5.907)$ \\
     & TVAE & $0.298\;(0.041)$ & $0.032\;(0.008)$ & $1.000\;(0.000)$ & $0.966\;(0.031)$ & $0.134\;(0.015)$ & $0.319\;(0.449)$ & $0.031\;(0.044)$ & $0.114\;(0.159)$ \\
    \midrule
    \multirow{4}{*}{\dataset{Steel}} & CTGAN & $0.032\;(0.003)$ & $0.032\;(0.003)$ & $0.993\;(0.005)$ & $0.669\;(0.114)$ & $0.647\;(0.028)$ & $0.007\;(0.009)$ & $0.000\;(0.000)$ & $0.003\;(0.005)$ \\
     & Gaussian Copula & $0.383\;(0.017)$ & $0.383\;(0.017)$ & $0.999\;(0.001)$ & $0.797\;(0.064)$ & $0.359\;(0.049)$ & $0.213\;(0.368)$ & $0.009\;(0.015)$ & $0.074\;(0.128)$ \\
     & TabDDPM & $0.000\;(0.000)$ & $0.000\;(0.000)$ & $0.949\;(0.031)$ & $0.503\;(0.021)$ & $0.951\;(0.003)$ & $0.003\;(0.004)$ & $0.000\;(0.000)$ & $0.000\;(0.001)$ \\
     & TVAE & $0.054\;(0.011)$ & $0.054\;(0.011)$ & $0.986\;(0.010)$ & $0.579\;(0.015)$ & $0.639\;(0.057)$ & $0.007\;(0.011)$ & $0.000\;(0.000)$ & $0.002\;(0.003)$ \\
    \midrule
    \multirow{4}{*}{\dataset{Taxi}} & CTGAN & $0.049\;(0.004)$ & $0.025\;(0.002)$ & $1.000\;(0.000)$ & $1.000\;(0.000)$ & $0.547\;(0.039)$ & -- & -- & -- \\
     & Gaussian Copula & $0.374\;(0.086)$ & $0.194\;(0.045)$ & $1.000\;(0.000)$ & $1.000\;(0.000)$ & $0.494\;(0.237)$ & -- & -- & -- \\
     & TabDDPM & $0.007\;(0.003)$ & $0.004\;(0.001)$ & $1.000\;(0.000)$ & $0.862\;(0.012)$ & $0.734\;(0.292)$ & -- & -- & -- \\
     & TVAE & $0.059\;(0.015)$ & $0.030\;(0.007)$ & $1.000\;(0.000)$ & $1.000\;(0.000)$ & $0.529\;(0.111)$ & -- & -- & -- \\
    \midrule
    \multirow{4}{*}{\dataset{URL}} & CTGAN & $0.033\;(0.012)$ & $0.017\;(0.006)$ & -- & -- & -- & $0.962\;(0.016)$ & $0.134\;(0.021)$ & $0.873\;(0.171)$ \\
     & Gaussian Copula & $0.025\;(0.001)$ & $0.012\;(0.001)$ & -- & -- & -- & $0.852\;(0.084)$ & $0.080\;(0.020)$ & $0.211\;(0.080)$ \\
     & TabDDPM & $0.020\;(0.035)$ & $0.010\;(0.018)$ & -- & -- & -- & $0.691\;(0.533)$ & $0.183\;(0.229)$ & $6.320\;(6.907)$ \\
     & TVAE & $0.009\;(0.001)$ & $0.005\;(0.000)$ & -- & -- & -- & $0.606\;(0.014)$ & $0.047\;(0.008)$ & $0.242\;(0.031)$ \\
    \bottomrule
  \end{tabular}
  }
  \caption{Raw constraint results for every dataset--generator combination.
  Entries are means, with standard deviations in parentheses, over nine runs
  spanning three independent data splits. CVR, sCVC, and LFD are lower-is-better
  metrics; equational $R^2$ is higher-is-better. Dashes indicate that a
  constraint family is unavailable for a dataset. News linear metrics are
  computed over the two splits containing detected linear constraints; the
  zero-constraint split is excluded ($n=2$).}
  \label{tab:full-raw-constraint-results}
\end{table*}

%% file: ablation_studies.tex
\clearpage

\section{Ablation Studies}
\label{app:ablation_studies}

\subsection{Constraint-Discovery Hyperparameter Sensitivity}
\label{app:discovery_hyperparameter_sensitivity}

We examine three inference-time hyperparameters: discovery rounds
$R_{\mathrm{disc}}$, context rows $n_{\mathrm{ctx}}$, and the maximum number of
validation counterexamples returned per revision $n_{\mathrm{cex}}$. Starting
from $(R_{\mathrm{disc}},n_{\mathrm{ctx}},n_{\mathrm{cex}})=(3,100,20)$, we
vary one parameter at a time, testing $R_{\mathrm{disc}}\in\{1,5\}$,
$n_{\mathrm{ctx}}\in\{50,200\}$, and $n_{\mathrm{cex}}\in\{0,50\}$.

We evaluate the default and the six one-factor variants above---seven
configurations in total---on three predefined splits of \dataset{News} with
GPT-5.6 Luna \citep{openai2026gpt56luna}, yielding 21 discovery runs. Each run
executes LD, equational, and linear discovery. We disable equational
repair-function generation (\texttt{fix(df)}) so that token usage isolates
constraint discovery.

For each run, \emph{retained-constraint yield} is the number of hypotheses
retained after full-table validation and family-specific consolidation, summed
across the three families before enforcement-specific cross-family pruning.
The count can therefore exceed its end-to-end counterpart. Because
\dataset{News} has no exhaustive gold constraint set, we treat higher yield as
desirable: a larger validated and consolidated hypothesis set provides evidence
of broader constraint coverage, rather than proof of exact semantic recovery.
Figure~\ref{fig:discovery-hyperparameter-sensitivity} visualizes both yield and
input-plus-output token usage, averaged over the three splits.

\textbf{Among the tested round counts, three discovery rounds best balance
yield and cost.} Reducing $R_{\mathrm{disc}}$ from three to one cuts token
use but lowers mean yield from 46.33 to 11.33. Increasing it to five more than
doubles token usage while yielding fewer retained constraints (38.67), showing
that additional rounds do not necessarily add unique hypotheses after
validation and consolidation.

\textbf{More context does not improve the cost--yield trade-off.} Relative
to the 100-row default, both 50 and 200 context rows consume more tokens while
producing similar yields (48.00 and 44.33 versus 46.33). Thus, 100 rows is the
most efficient context budget among those tested.

\textbf{Counterexample feedback strongly improves retained-constraint
yield.} Removing it reduces mean yield from 46.33 to 19.33 without lowering
token usage. Raising $n_{\mathrm{cex}}$ from 20 to 50 produces a similar yield
(47.33) at greater cost. 

Together, these results support $(3,100,20)$ as an
efficient operating point in this setting; this one-factor-at-a-time study on
one dataset and backbone does not establish a universally optimal configuration
or statistical significance.

\subsection{Equational Repair Order and Target Selection}
\label{app:equational_repair_order_ablation}

To isolate the effect of equational repair order and target selection, we
reuse the synthetic tables and discovered constraints from the end-to-end
experiment in Section~\ref{sec:end_to_end_evaluation}. We rerun only the
equational-repair stage; LD and linear-inequality enforcement are disabled.
The synthetic data, splits, generators, constraints, and validated
target-specific repair functions are held fixed. The only intervention is the
scheduling policy: our KS-complement-guided method
(Appendix~\ref{app:equational_repair}) versus a structurally feasible random
schedule.

Once an equation is repaired, all of its participating columns are frozen
as future targets so that later reconstructions cannot invalidate it. Among
all complete repair schedules satisfying this dependency rule, the KS-guided
method first selects those with the highest minimum predicted change in KS
complement and then chooses the schedule with the largest total predicted
change. The random baseline instead randomly selects repair targets and their
order, using backtracking whenever a choice prevents completion of the
schedule.

The evaluation covers the five datasets containing discovered equational
constraints---\dataset{Flights}, \dataset{NBA}, \dataset{News},
\dataset{Taxi}, and \dataset{Steel}---with four generators, three data splits,
and three synthetic samples per split, yielding 36 matched inputs per
dataset and 180 in total. Scheduling becomes consequential when equations
share columns or admit multiple validated target-specific repairs, because
each resolved equation removes columns from later target choices. We evaluate
univariate marginal fidelity using Column Shapes and constraint
satisfaction using equational CVR. Table~\ref{tab:equational-repair-order-aggregated}
reports dataset-level results aggregated across generators, splits, and
synthetic samples.

\input{equational_repair_order_aggregated}

{\textbf{KS guidance yields higher marginal fidelity while both schedulers
achieve exact equation satisfaction.} Both feasible schedules produce zero
equational CVR on all 180 matched inputs, so their difference lies in fidelity
rather than constraint satisfaction. Relative to random feasible scheduling,
KS guidance improves mean paired Column Shapes on four datasets by
$0.020$--$0.053$, with the largest gain on \dataset{Flights}, and is effectively
tied on \dataset{News} at the reported precision ($0.000$). The equal-weight
mean improvement across datasets is $0.025$. Thus, in these settings, KS
guidance yields higher univariate marginal fidelity than random feasible
selection, without claiming global optimality.}

\subsection{Equational--Linear Repair Order}
\label{app:cross_family_repair_order_ablation}

We next ablate the order of the two numerical repair stages. Logical
dependencies involve only categorical columns, so their participating columns
are disjoint from those involved in equational and linear repair. Consequently,
LD repair cannot interfere with either numerical constraint family, and we
apply it first in both conditions. We then vary the numerical order:
\emph{equational then linear} ($E{\rightarrow}L$), our default, versus
\emph{linear then equational} ($L{\rightarrow}E$).

This order matters because equations and linear inequalities can share
numerical columns. Under $E{\rightarrow}L$, equational repair first establishes
the equations; the subsequent projection then protects equation-participating
columns and adjusts only the remaining mutable columns, thereby preserving the
repaired equations while enforcing the inequalities. Under
$L{\rightarrow}E$, no equations have yet been repaired, so the projection has
no equation-derived protected columns and may modify any participating
numerical column. Subsequent equational repair can then change shared columns
again and move previously projected rows outside the linear feasible region.

We focus on \dataset{NBA} because its mined equational and linear constraints
share participating columns. We reuse the trained generators, mined
constraints, and corresponding synthetic tables from the end-to-end
experiment. Both conditions start from the same synthetic table and hold
the discovered constraints and family-specific repair routines fixed; the only
intervention is the order of the two numerical stages. The ablation spans
three data splits, four generators, and three
independent synthetic draws per split, yielding 36 matched
generator--split--draw settings (nine per generator).
Table~\ref{tab:cross-family-repair-order} reports the final linear CVR and LFD,
aggregated over the nine settings for each generator. We also verify the
final equational CVR under both orders to detect cross-family regressions.

\begin{table}[t]
    \centering
    \scriptsize
    \setlength{\tabcolsep}{3pt}
    \begin{tabular}{@{}llrr@{}}
        \toprule
        Generator & Metric & $E{\rightarrow}L$ & $L{\rightarrow}E$ \\
        \midrule
        \multirow{2}{*}{CTGAN}
            & CVR (\%) $\downarrow$ & $0\;(0)$ & $7.331\;(5.919)$ \\
            & LFD $\downarrow$ & $0\;(0)$ & $1.583\;(2.102)$ \\
        \addlinespace[1pt]
        \multirow{2}{*}{TVAE}
            & CVR (\%) $\downarrow$ & $0\;(0)$ & $0.782\;(1.193)$ \\
            & LFD $\downarrow$ & $0\;(0)$ & $0.640\;(1.225)$ \\
        \addlinespace[1pt]
        \multirow{2}{*}{Gaussian Copula}
            & CVR (\%) $\downarrow$ & $0\;(0)$ & $9.919\;(3.753)$ \\
            & LFD $\downarrow$ & $0\;(0)$ & $1.219\;(0.755)$ \\
        \addlinespace[1pt]
        \multirow{2}{*}{TabDDPM}
            & CVR (\%) $\downarrow$ & $0\;(0)$ & $0.262\;(0.366)$ \\
            & LFD $\downarrow$ & $0\;(0)$ & $0.012\;(0.014)$ \\
        \midrule
        \multirow{2}{*}{\textbf{All}}
            & CVR (\%) $\downarrow$ & $\mathbf{0\;(0)}$ &
            $\mathbf{4.574\;(5.419)}$ \\
            & LFD $\downarrow$ & $\mathbf{0\;(0)}$ &
            $\mathbf{0.863\;(1.360)}$ \\
        \bottomrule
    \end{tabular}
    \caption{Repair-order ablation on \dataset{NBA}. $E{\rightarrow}L$ is
    categorical $\rightarrow$ equational $\rightarrow$ linear repair;
    $L{\rightarrow}E$ swaps the two numerical stages. The \textbf{All} row
    aggregates all 36 matched settings. Entries are means, with sample standard
    deviations in parentheses. Final equational CVR is zero under both
    orders and is therefore omitted.}
    \label{tab:cross-family-repair-order}
\end{table}

\textbf{Only $E{\rightarrow}L$ achieves joint satisfaction in all 36
matched settings.} Its final linear CVR and LFD are both zero, while the
equations remain exactly satisfied. In contrast,
$L{\rightarrow}E$ ends with linear violations in 24 of 36 settings ($66.7\%$),
raising aggregate CVR to $4.574\%$ and LFD to $0.863$. The reversed order yields
nonzero mean CVR for every generator, led by Gaussian Copula ($9.919\%$) and
CTGAN ($7.331\%$). Because both orders retain zero equational CVR, it loses
linear feasibility rather than trading satisfaction between the two families.

\textbf{Repair interference is directional.}
Under $L{\rightarrow}E$, equational reconstruction changes shared columns
after projection and can move rows outside the linear feasible region. Under
$E{\rightarrow}L$, the final projection protects equation-participating
columns and, in these settings, restores linear feasibility without undoing
the equations. Thus, projection last is the safer order when the two families
overlap. Because \dataset{NBA} was selected specifically for such overlap,
this ablation tests that interference mechanism rather than claiming that
order matters when the families use disjoint columns.

%% file: equational_repair_order_aggregated.tex
\begin{table}[t]
    \centering
    \small
    \setlength{\tabcolsep}{3pt}
    \begin{tabular}{@{}lrrr@{}}
        \toprule
        & \multicolumn{3}{c}{Column Shapes $\uparrow$} \\
        \cmidrule(l){2-4}
        Dataset
            & KS-guided
            & \shortstack{Random\\feasible}
            & KS gain \\
        \midrule
        \dataset{Flights}
            & $0.878(0.037)$
            & $0.825(0.048)$
            & $\mathbf{+0.053(0.020)}$ \\
        \dataset{NBA}
            & $0.916(0.026)$
            & $0.897(0.040)$
            & $\mathbf{+0.020(0.024)}$ \\
        \dataset{News}
            & $0.819(0.102)$
            & $0.818(0.101)$
            & $0.000(0.001)$ \\
        \dataset{Taxi}
            & $0.898(0.043)$
            & $0.875(0.051)$
            & $\mathbf{+0.023(0.015)}$ \\
        \dataset{Steel}
            & $0.890(0.046)$
            & $0.863(0.060)$
            & $\mathbf{+0.028(0.031)}$ \\
        \bottomrule
    \end{tabular}
    \caption{Equational-repair scheduling ablation, aggregated by dataset
    over four generators, three data splits, and three synthetic samples per
    split (36 matched inputs per dataset; 180 total). Values are means with
    sample standard deviations in parentheses. KS gain is computed per
    matched input as KS-guided minus random feasible, so positive values favor
    KS guidance. Equational CVR is zero under both schedules on every input and
    is therefore omitted.}
    \label{tab:equational-repair-order-aggregated}
\end{table}